\documentclass[10pt,twocolumn,letterpaper]{article}

\usepackage{cvpr}              % To produce the CAMERA-READY version
\definecolor{cvprblue}{rgb}{0.21,0.49,0.74}
\usepackage[pagebackref,breaklinks,colorlinks,allcolors=cvprblue]{hyperref}
\usepackage{algorithm}
\usepackage{algpseudocodex}
\usepackage{amsmath}
\usepackage{amssymb}
\usepackage{multirow}
\usepackage{booktabs}
\usepackage{makecell}
\usepackage{subcaption}
\usepackage{fancybox}
\usepackage{fancyvrb}
\usepackage{bm}
\usepackage{multirow}
\usepackage{booktabs}
\usepackage{bbding}
\usepackage{makecell}
\usepackage{subcaption}
\usepackage{fancybox}
\usepackage{fancyvrb}
\usepackage{colortbl}
\usepackage{xcolor}
\usepackage{url}

\usepackage[accsupp]{axessibility}  % Improves PDF readability for those with disabilities.

\definecolor{softred}{RGB}{255, 178, 178}  % Softer red for highest values
\definecolor{softorange}{RGB}{255, 218, 179} % Softer orange for second highest values
\definecolor{softyellow}{RGB}{255, 244, 191} % Softer yellow for third highest values

\def\paperID{7692} % *** Enter the Paper ID here
\def\confName{CVPR}
\def\confYear{2025}

\title{ReasonGrounder: LVLM-Guided Hierarchical Feature Splatting for \\
Open-Vocabulary 3D Visual Grounding and Reasoning}

\author{Zhenyang Liu$^{1,3}$, Yikai Wang$^{2}$\footnotemark[1], Sixiao Zheng$^{1,3}$, Tongying Pan$^{1}$, \\Longfei Liang$^{4}$, Yanwei Fu$^{1,3}$\footnotemark[2], Xiangyang Xue$^{1}$\footnotemark[2]\\
$^1$\normalsize Fudan University
$^2$\normalsize Nanyang Technological University
$^3$\normalsize Shanghai Innovation Institute $^4$\normalsize NeuHelium Co., Ltd\\
{\tt\small lzyzjhz@163.com, yikai.wang@ntu.edu.sg, sxzheng18@fudan.edu.cn, typan23@m.fudan.edu.cn,} \\
{\tt\small longfei.liang@neuhelium.com,
\{yanweifu,xyxue\}@fudan.edu.cn}\\
Project Page: \href{https://zhenyangliu.github.io/ReasonGrounder/}{ZhenyangLiu.github.io/ReasonGrounder}}
\begin{document}
\maketitle
\renewcommand{\thefootnote}{\fnsymbol{footnote}}
\footnotetext[1]{The work was completed while Yikai was at Fudan.}
\footnotetext[2]{Corresponding authors.}
\footnotetext[3]{Prof. Yanwei Fu is also with Institute of Trustworthy Embodied Al, and the School of Data Science, Fudan University.}
\begin{abstract}
Open-vocabulary 3D visual grounding and reasoning aim to localize objects in a scene based on implicit language descriptions, even when they are occluded. This ability is crucial for tasks such as vision-language navigation and autonomous robotics. However, current methods struggle because they rely heavily on fine-tuning with 3D annotations and mask proposals, which limits their ability to handle diverse semantics and common knowledge required for effective reasoning. In this work, we propose ReasonGrounder, an LVLM-guided framework that uses hierarchical 3D feature Gaussian fields for adaptive grouping based on physical scale, enabling open-vocabulary 3D grounding and reasoning. ReasonGrounder interprets implicit instructions using large vision-language models (LVLM) and localizes occluded objects through 3D Gaussian splatting.
By incorporating 2D segmentation masks from the SAM and multi-view CLIP embeddings, ReasonGrounder selects Gaussian groups based on object scale, enabling accurate localization through both explicit and implicit language understanding, even in novel, occluded views. We also contribute ReasoningGD, a new dataset containing over 10K scenes and 2 million annotations for evaluating open-vocabulary 3D grounding and amodal perception under occlusion. Experiments show that ReasonGrounder significantly improves 3D grounding accuracy in real-world scenarios.
\end{abstract}

\section{Introduction}
\label{sec:intro}

\begin{figure}[t]
	\centering
	\includegraphics[width=1\linewidth]{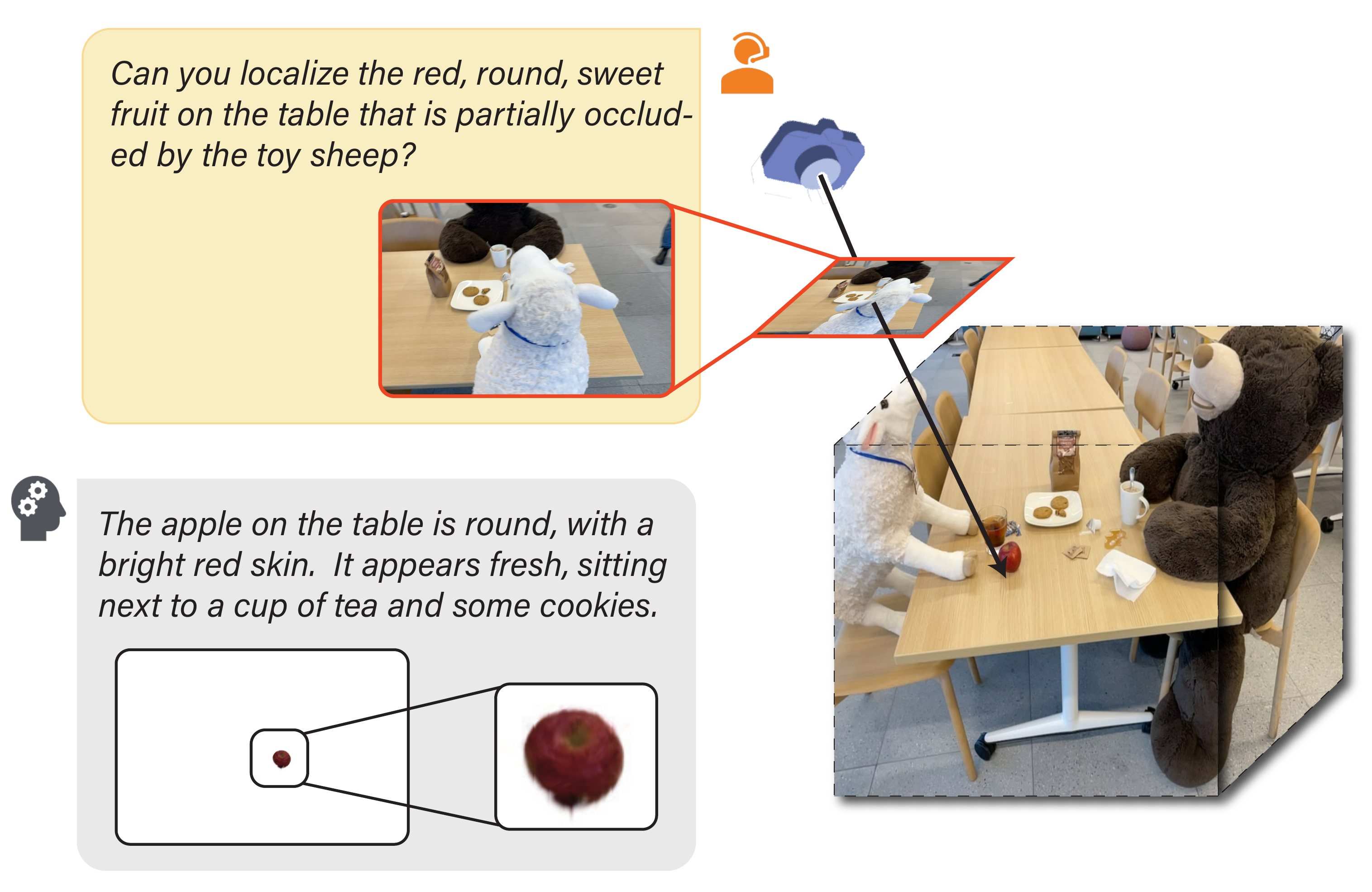}
        \vspace{-0.2in}
	\caption{\textbf{Examples of open-vocabulary 3D visual grounding and reasoning.}\label{teaser} In a given scene, the user observes from a perspective with occlusions and asks questions such as: ``Can you localize the red, round, sweet fruit on the table that is partially occluded by the toy sheep?" Open-vocabulary 3D visual grounding and reasoning seeks to interpret complex implicit queries, deduce answers, and accurately localize the target object, even when it is partially or fully occluded from the current viewpoint.}
\vspace{-0.25in}
\end{figure}

Open-vocabulary 3D visual grounding and reasoning aims to accurately localize objects in a scene using implicit language descriptions, even when the objects are occluded or only partially visible. This task is challenging as language complexity requires both comprehension and advanced visual reasoning. For instance, simple commands like \textit{apple} can be directly interpreted, while more complex instructions, such as \textit{Can you localize the red, round, sweet fruit on the table that is partially occluded by the toy sheep?}, require more from the system. As shown in Figure~\ref{teaser}, these instructions involve locating the target object, interpreting the intent of the description, and accounting for occlusion.
This capability is crucial for applications such as autonomous robotics~\cite{vemprala2024chatgpt,firoozi2023foundation} and augmented reality~\cite{arena2022overview,carmigniani2011augmented}, where systems must handle dynamic, unstructured, and often incomplete visual data while interpreting vague or indirect language to infer user intent and complete complex tasks.

Existing 3D visual grounding (3DVG) methods~\cite{huang2022multi,guo2023viewrefer,yang2024llm,chen2022language} face challenges in open-vocabulary grounding and reasoning, primarily due to reliance on 3D annotations~\cite{zhu2024scanreason, yang2021sat} and mask proposals~\cite{chen2020scanrefer, achlioptas2020referit3d}, which limit generalization across dynamic environments. Despite its importance for vision-language navigation and robotics, 3DVG struggles with the complexities of real-world scenes, where objects and layouts constantly change. Recent methods~\cite{kerr2023lerf,qin2024langsplat} aim to localize objects using open-vocabulary queries without 3D annotations, leveraging 3D radiance fields (e.g., NeRF~\cite{mildenhall2021nerf}, 3DGS~\cite{kerbl20233d}) and 2D models like CLIP. However, challenges remain in interpreting user intent and handling occlusions during object localization.

To achieve open-vocabulary 3D visual grounding and reasoning, this paper proposes ReasonGrounder, a novel LVLM-Guided Hierarchical Feature Splatting method that enables implicit instruction comprehension and amodal perception of novel views. ReasonGrounder uses hierarchical 3D Gaussian feature fields to enhance 3D Gaussian features, enabling adaptive Gaussian grouping based on physical scale and precise open-vocabulary 3D visual grounding. Inspired by the remarkable comprehension and reasoning capabilities of large vision-language models (LVLM), ReasonGrounder leverages the common knowledge of LVLM to interpret implicit instructions and reason about the underlying intention and target object. Based on the reasoned intention, the hierarchical 3D feature Gaussians are grouped according to the physical scale of the target object. ReasonGrounder selects the hierarchical Gaussian group to localize the complete target object, even when the object is partially visible or completely occluded in novel views.

\begin{figure*}
	\centering
	\includegraphics[width=\linewidth]{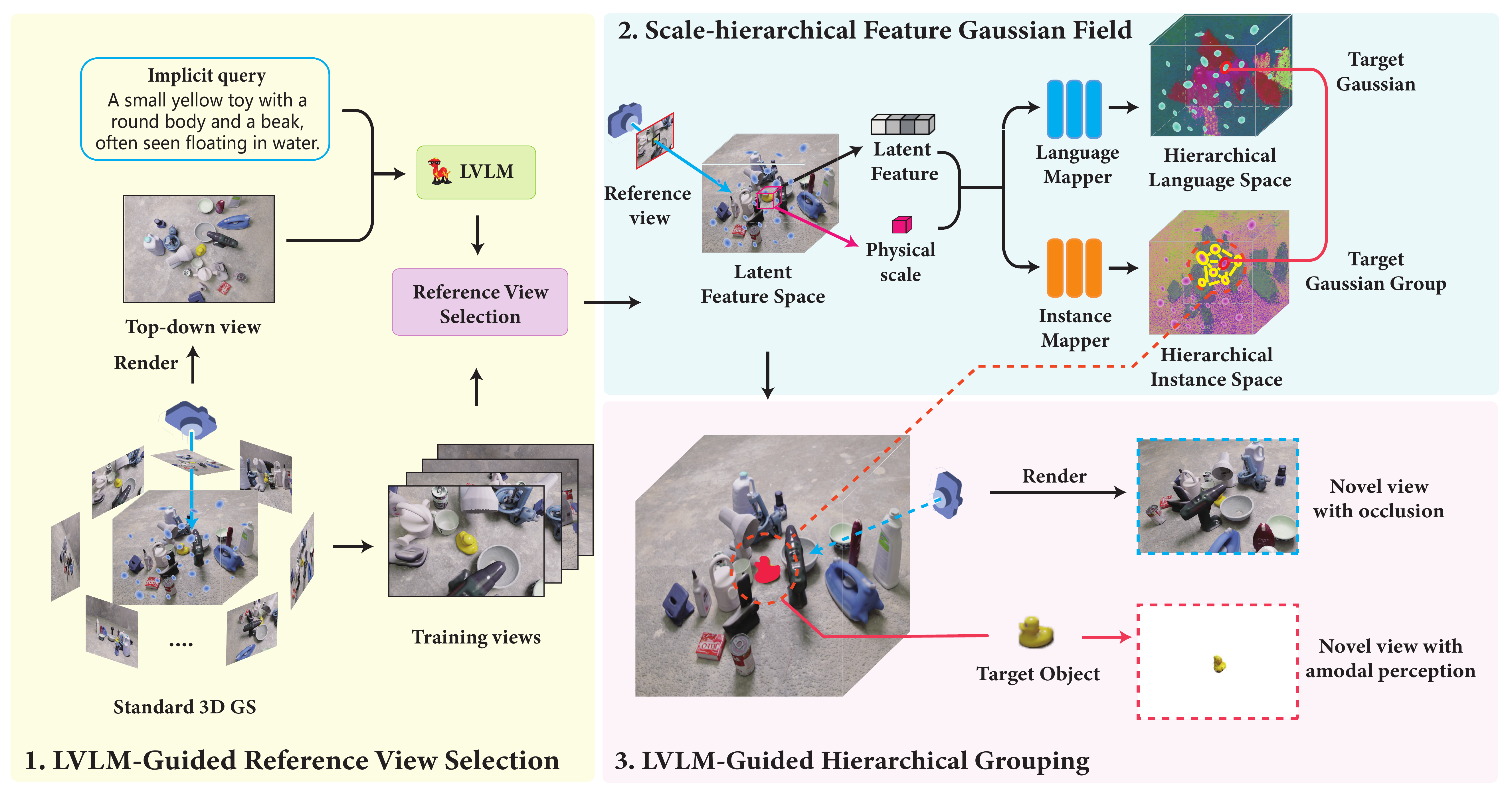}
        \vspace{-0.3in}
	\caption{\textbf{The framework of our ReasonGrounder.} Our \textbf{ReasonGrounder} leverages 3D Gaussian Splatting (3DGS) for efficient high-resolution rendering. It extracts 2D segmentation masks from SAM~\cite{kirillov2023segment} and maps them into a 3D field. Each mask is assigned a 3D scale based on depth from 3DGS. Latent feature vectors are appended to Gaussians and mapped into hierarchical language and instance features using two MLPs. CLIP embeddings ensure multi-view consistency, while contrastive loss refines the masks. A reference view is selected based on LVLM, guiding implicit instruction comprehension for accurate 3D localization and amodal perception in novel views.\label{fig:overall} }	
	 \vspace{-0.2in}
\end{figure*}

Formally, we propose the ReasonGrounder framework as illustrated in Fig.~\ref{fig:overall}. ReasonGrounder employs 3D Gaussian Splatting (3DGS), which represents scenes as 3D Gaussian collections with tile-based splatting for efficient, high-resolution rendering. In particular, a standard 3DGS scene is constructed, and 2D segmentation masks from SAM~\cite{kirillov2023segment} are projected into a 3D field. For each mask, a 3D scale is calculated from the depth rendered by the 3DGS. To enhance each Gaussian’s view-independent representation, ReasonGrounder appends a latent feature vector mapped into hierarchical language and instance features via two shallow MLPs: a language mapper and an instance mapper. CLIP embeddings supervise language features across views for multi-view consistency, while instance features refine 2D mask candidates using contrastive loss and 3D scale, supporting feature-based Gaussian grouping.
Further, to aid localization, an instruction-conditioned mechanism guided by LVLM selects the reference view most aligned with the instruction. This view and instruction enable comprehension of the target object of intent. Using 3D scale and hierarchical feature Gaussians, ReasonGrounder achieves precise 3D localization and amodal perception in novel views.

Furthermore, we introduce a novel \textit{ReasoningGD} dataset containing over 10K complex scenes and 263 object types, with a total of approximately 2 million annotations. Each scene, generated using Blenderproc~\cite{denninger2019blenderproc}, includes point clouds, 100 RGB-D images with detailed labels, camera poses, and 2D modal/amodal masks. The dataset features a variety of object instances with varying levels of occlusion, enabling comprehensive evaluation of open-vocabulary 3D visual grounding and reasoning, including 3D localization with implicit instructions and amodal perception of views.

The contributions of this paper are summarized:
\begin{itemize}
    \item \textbf{(1) \textit{Enhanced 3D Visual Grounding}}: We address limitations in open-vocabulary 3D visual grounding, such as reliance on 3D annotations and limited semantic understanding, by using hierarchical 3D Gaussian fields with LVLMs for robust grounding and reasoning.
    \item \textbf{(2) \textit{A Novel ReasonGrounder Framework}}: The proposed ReasonGrounder leverages hierarchical 3D feature Gaussian fields for adaptive Gaussian grouping with 3D scale, enabling effective open-vocabulary 3D visual grounding and reasoning. ReasonGrounder interprets implicit instructions using large vision-language models (LVLM) and accurately localizes occluded objects with hierarchical 3D feature Gaussian splatting.
    \item \textbf{(3) \textit{Hierarchical Feature Splatting and Amodal Perception}}: ReasonGrounder empowers the hierarchical features of 3D Gaussians and selects Gaussian groups based on the target object's scale. LVLM aids in interpreting complex instructions and locating objects even when partially or fully occluded.
    \item \textbf{(4) \textit{Dataset Contributions}}: A new ReasoningGD dataset offers over 10K complex scenes with 2 million annotations, including point clouds, RGB-D images with detailed labels, camera poses, and 2D modal/amodal masks of views, enabling rigorous evaluation of 3D visual grounding with implicit instruction handling and occlusion robustness.
\end{itemize}

% \end{itemize}

% 3D visual grounding aims to localize the referenced object within a 3D scene using language expressions. To address this task, ScanRefer~\cite{chen2020scanrefer} and ReferIt3D~\cite{achlioptas2020referit3d} first created datasets. Earlier works employed a two-stage pipeline using pre-trained detectors to generate object recommendations and extract features, such as PointNet++~\cite{DBLP:journals/corr/QiYSG17}. InstanceRefer~\cite{DBLP:journals/corr/abs-2103-01128} transforms this task into an instance-matching problem.	Additionally, some studies~\cite{chen2022unit3dunifiedtransformer3d,jin2023context} have investigated 3D language pretraining using advanced techniques such as mask modeling and contrastive learning on paired object-caption data, followed by fine-tuning for downstream tasks. However, it still requires extensive data annotations to train the neuro-symbolic networks, limiting its open-vocabulary zero-shot capabilities.

\section{Related Work}
\noindent\textbf{3D Visual Grounding.} 
3D visual grounding localizes objects in a scene based on language input. ScanRefer~\cite{chen2020scanrefer} and ReferIt3D~\cite{achlioptas2020referit3d} pioneered this task. Early approaches used two-stage pipelines with pre-trained object detectors, such as PointNet++\cite{qi2017pointnet++}, for object proposals and feature extraction. InstanceRefer\cite{yuan2021instancerefer} reformulated the task as instance matching. More recent works~\cite{chen2022unit3dunifiedtransformer3d, jin2023context} have explored 3D language pretraining with mask modeling and contrastive learning. However, these methods still rely on extensive annotations, which limits their zero-shot performance.

\noindent\textbf{3D Language Fields.} 
3D language fields are emerging to learn 3D-consistent features by extending 2D model outputs across views. Methods like \cite{li2024place} and \cite{Siddiqui_2023_CVPR} enhance 2D semantics and segmentation for open-vocabulary 3D grounding. Early approaches include Distilled Feature Fields~\cite{kobayashi2022decomposingnerfeditingfeature} and Neural Feature Fusion Fields~\cite{tschernezki22neural}. LERF~\cite{kerr2023lerf} integrated CLIP features into NeRF for open-vocabulary 3D queries. To address speed and accuracy issues, LangSplat~\cite{qin2024langsplat} introduced a 3D Gaussian Splatting-based method. Despite progress, these methods rely on explicit instructions and struggle with occlusions.

\noindent\textbf{Large Vision Language Models.} Inspired by LLMs' reasoning abilities, researchers developed LVLMs to bring these skills to vision. Models like BLIP-2~\cite{li2023blip}, LLaVA~\cite{liu2024visual}, and MiniGPT-4~\cite{zhu2023minigpt} align visual and linguistic embeddings via large image-text datasets. LISA~\cite{lai2024lisa} integrates segmentation for better reasoning, while ScanReason~\cite{zhu2024scanreason} applies LVLMs to 3D but remains closed-vocabulary. Reasoning3D~\cite{chen2024reasoning3d} supports open-vocabulary segmentation but depends on predefined 3D models and lacks scenario-level reasoning. In contrast, our ReasonGrounder enables open-vocabulary 3D reasoning and grounding without annotations or predefined models, supporting implicit instructions and amodal perception.

\section{Our Proposed ReasonGrounder}
As shown in Figure~\ref{fig:overall}, we introduce ReasonGrounder, an LVLM-guided framework that uses scale-hierarchical feature Gaussian fields for adaptive 3D scale-based grouping. 

\begin{figure*}
	\centering
	\includegraphics[width=1\linewidth]{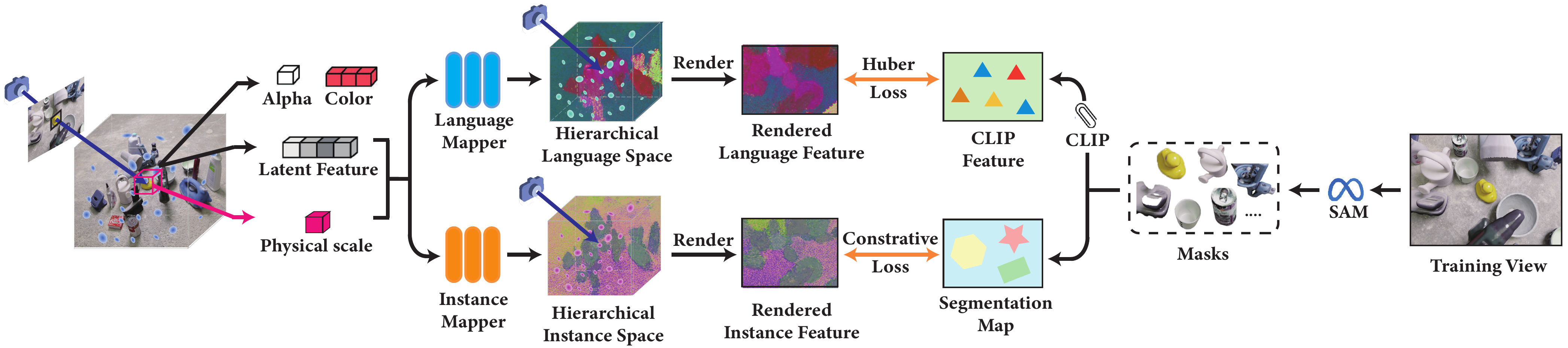}
    \vspace{-0.2in}
	\caption{\textbf{The pipeline of scale-hierarchical feature Gaussian field.} The method extracts 2D masks from SAM and projects them into a 3D field. ReasonGrounder adds a latent feature to each Gaussian, mapping it into hierarchical language and instance features. Language features are supervised by CLIP embeddings, while instance features refine masks using contrastive loss and 3D scale.\label{fig:train}}
    \vspace{-0.15in}
\end{figure*}

\subsection{Scale-Hierarchical Feature Gaussian Field}\label{ms1}
ReasonGrounder first trains a standard 3D Gaussian Splatting field for 3D representation. Pre-trained 2D model features serve as supervision for the scale-hierarchical Gaussian field using differentiable rasterization. This process forms a 3D feature Gaussian Splatting, with a latent feature appended to each Gaussian. The latent feature is mapped with physical scale into hierarchical features, comprising language and instance features.

\noindent\textbf{Supervision Generation.}
To train ReasonGrounder, input images are initially pre-processed using the Segment Anything Model (SAM) to generate precise object masks from training views. These masks are then used to assign physical scales and extract pixel-aligned features. Specifically, ReasonGrounder pretrains a standard 3D Gaussian field and employs SAM's automatic mask generator to create segmentation masks from training images. These masks are then filtered based on confidence, and nearly identical ones are deduplicated to generate final mask candidates $\{m_1, m_2, \dots, m_n\}$. To calculate the physical scales of masks, we obtain the depth of pixels for each mask based on the trained 3D Gaussian field. Given the ray origin, the pixels of masks can be deprojected onto 3D points with the focal length and depth. By calculating the standard deviation of these points, we can estimate the physical scales $\{s_1, s_2, \dots, s_n\}$.
With the obtained mask candidates, we proceed to extract CLIP features $\{\phi_1, \phi_2, \dots, \phi_n\}$ for each segmented region across multiple training views, thereby ensuring multi-view consistency. Thus, for each mask candidate, we obtain the associated triplet $\{m_i, \phi_i, s_i\}$.

\noindent\textbf{3D Feature Gaussian Splatting.} 
3D Gaussian Splatting directly models a 3D environment using a set of anisotropic 3D Gaussian functions. Each Gaussian $G(x)$ at position $x$ is defined by a mean $\mu \in \mathbb{R}^3$ and a covariance matrix $\Sigma$: 
\begin{equation}
	G(x) = \exp (- \frac{1}{2} (x - \mu)^{\top} \Sigma^{-1} (x -  \mu)).
	\label{eq:3dGaussian}
\end{equation}
To optimize the parameters of 3D Gaussians, they are rendered into 2D image planes, and a tile-based rasterizer is used to improve the rendering efficiency:
\begin{equation}
	C = \sum_{i \in \mathcal{N}} c_i \alpha_i \prod_{j=1}^{i-1} (1 - \alpha_j),
	\label{eq:rendering_3dgs}
\end{equation}
where $c_i$ is the color of the $i$-th Gaussian, $\mathcal{N}$ denotes the Gaussians in the tile, $C$ is the rendered color, and $\alpha_i = o_i \sigma_i$. Here, $o_i$ is the opacity of the $i$-th Gaussian and $\sigma_i$ represents the function of the $i$-th Gaussian projected onto 2D.

In this paper, we extend Gaussian Splatting by embedding latent features $f_{g_i}$ within each 3D Gaussian, where $f_{g_i} \in \mathbb{R}^{d_f}$ and $d_f$ indicates its dimension. Similar to Eqn.~\ref{eq:rendering_3dgs}, we employ differential Gaussian rasterization to obtain latent features for each 2D image $I$, denoted as $\bar{F} = \{\bar{f_i}\}$, where $i$ denotes each pixel of $I$:
\begin{equation}
	\bar{f_i} = \sum_{i \in \mathcal{N}} f_{g_i} \alpha_i \prod_{j=1}^{i-1} (1 - \alpha_j),
	\label{eq:rendering_f3dgs}
\end{equation}
Subsequently, the latent feature is mapped into hierarchical features, comprising language and instance features.

\noindent\textbf{Hierarchical Language Feature.} Having obtained the language embedding $\{\phi_i\}$ with 3D physical scale $\{s_i\}$ on 2D posed images, we can learn a 3D hierarchical language feature modeling the relations between 3D points and 2D pixels. However, directly incorporating high-dimensional language embeddings into Gaussians would lead to significant memory overhead. ReasonGrounder employs principal component analysis (PCA)~\cite{abdi2010principal} to compress the features. The compressed features $\{\hat{\phi_i}\}$ remain sufficient for scene representation, and the memory overhead is significantly reduced. ReasonGrounder adopts a shallow MLP as the language mapper $F_{l}$, taking the 3D scale $s_i$ and latent feature $f_{g_i}$ as inputs:
\begin{equation}
	\phi_{g_i}^s = F_{l}(s_i, f_{g_i})
	\label{eq:langrendering_f3dgs}
\end{equation}
$\phi_{g_i}^s$ denotes the hierarchical language feature. Since all Gaussian latent features share a common language mapper $F_{l}$ at scale $s_i$ during training, we first render the latent features into 2D, denoted as $\bar{f_i}$, and then apply $F_{l}$ to these 2D features to derive $\phi_i^s$, as shown in Eqn.~\ref{eq:langrendering_f3dgs}. The loss function $L_{lang}$ is defined as follows:
\begin{equation}
	L_{lang} = L_{\delta}(\phi_i^s, \hat{\phi_i})
	\label{loss_lang}
\end{equation}
$L_{\delta}$ is the Huber loss with hyperparameter $\delta$. Hierarchical language features enable accurate 3D visual grounding.

\noindent\textbf{Hierarchical Instance Feature.}
An object in the 3D Gaussian Splatting field is essentially a group of Gaussians. Hierarchical instance features facilitate multi-level granularity Gaussian grouping within the scene, supporting precise object localization. Given a Gaussian latent feature $f_{g_i}$ and 3D scale $s_i$, the scale-hierarchical feature Gaussian field introduces another shallow MLP as the instance mapper $F_g$:
\begin{equation}
	\psi_{g_i}^s = F_{g}(s_i, f_{g_i})
	\label{eq:georendering_f3dgs}
\end{equation}
where $\psi_{g_i}^s$ denotes the 3D instance embedding of the Gaussian. These embeddings can be rendered using the same alpha blending as in Eqn.~\ref{eq:rendering_f3dgs} to obtain the pixel instance feature $\hat{\psi}^{s_i}_k$ corresponding to the ray $k$. The pixel instance feature reflects the 2D mask that a pixel belongs to at specific scales. Intuitively, if two pixels share at least one mask at a given scale $S_i$, they should have similar instance features.
We adapt the contrastive supervision loss proposed by GARField~\cite{kim2024garfield} for training the instance hierarchical space. Specifically, two rays $r_m$ and $r_n$ are cast from the pixels in the masks $M_m$ and $M_n$ with physical scales $S_m$ and $S_n$. The instance features $\hat{\psi}^{s_m}_m$ and $\hat{\psi}^{s_n}_n$ are volumetrically rendered along the cast rays. The concrete contrastive loss $L_{in}$ is detailed as follows:
\begin{equation}\label{eq4}
	L_{in} = \left\{ 
	\begin{aligned}
		& \lVert \hat{\psi}^{s_m}_m - \hat{\psi}^{s_n}_n \rVert  &\quad M_m = M_n\\
		& \rm ReLU(\lambda - \lVert \hat{\psi}^{s_m}_m - \hat{\psi}^{s_n}_n \rVert) & \quad  M_m \neq M_n\\
	\end{aligned}\right.
\end{equation}
where $\lambda$ denotes the bound constant. Note that this loss is employed among rays sampled from the same viewpoint. This supervision method enables multiple fine-grained Gaussian groupings, facilitating flexible and accurate localization of target objects.	

 \subsection{LVLM-Guided Hierarchical Grouping}\label{ms2}
Unlike current open-vocabulary 3D visual grounding methods like LERF~\cite{kerr2023lerf} and LangSplat~\cite{qin2024langsplat}, which handle explicit prompts, ReasonGrounder emphasizes implicit instruction comprehension and amodal perception of views.

\noindent \textbf{LVLM-Guided Reference View Selection.} Existing methods~\cite{delitzas2023multi,parelli2023clip} incorporate 2D information from top-down views, but this approach cannot handle our task: Top-down images are rarely used in 2D vision models, and their complexity and incompleteness hinder effective integration into standard vision-language models. Our ReasonGrounder introduces a query-conditioned 2D view selection mechanism that addresses this challenge by selecting 2D views most semantically related to the queries. We first input the top-down view $I_{td}$ and implicit query $Q_{im}$ into the LVLM $F_{vl}$ to reason the target object $O_t$ implied in the query, along with the corresponding explanation $E$:
\begin{equation}
	O_t, E = F_{vl}(I_{td}, Q_{im})
	\label{eq:rearendering_f3dgs}
\end{equation}
Then we utilize CLIP’s image-text encoder ($E_{im}$ and $E_{te}$) to select the most suitable 2D view as the reference view from training views $\{V_i\}$, guided by the target object.
\begin{equation}
	\hat{V} = \arg\max_{V_i} \frac{E_{im}(V_i) \cdot E_{te}(O_t)}{\|E_{im}(V_i)\| \|E_{te}(O_t)\|}
	\label{eq:rerendering_f3dgs}
\end{equation}
where $\hat{V}$ denotes the selected reference view.

\noindent \textbf{LVLM-Guided Hierarchical Gaussian Grouping.} To accurate localize complete objects in the presence of occlusions, our ReasonGrounder introduces hierarchical feature Gaussian grouping to efficiently address this limitation. Specifically, we first compute the CLIP embedding $\phi_{o_t}$ of the target object $O_t$, along with a set of canonical phrases $\phi_\text{canon}^m$. To assign the rendered language embedding $\phi_\text{lang}^i$ of pixel $i$ in the reference view $\hat{V}$ a score, we compute the relevancy $r_i$ between the rendered embedding and the canonical phrase embeddings: $r_i=\text{min}_m~\frac{\text{exp}(\phi_\text{lang}^i\cdot\phi_\text{quer})}{\text{exp}(\phi_\text{lang}^i\cdot\phi_\text{canon}^m)+\text{exp}(\phi_\text{lang}^i\cdot\phi_\text{quer}))}$
All renderings use the same canonical phrases: \textit{``object"}, \textit{``things"}, \textit{``stuff"}, and \textit{``texture"}. The selected reference view has obtained mask candidates, language embeddings, and 3D physical scales as $\{m_i, \phi_i, s_i\}$ during supervision generation, and ReasonGrounder selects the 3D scale $s_{i^*}$ corresponding to $\phi_i$ that is most relevant to the CLIP embedding $\phi_{o_t}$:
$s_{i^*}\hspace{-0.2em}=\hspace{-0.2em}\arg\max_{s_i} \text{sim}(\phi_{o_t}, \phi_i)$, where $\text{sim}$ denotes the cosine similarity. Then $s_{i^*}$ is used to hierarchize the feature Gaussian field $\psi_{g_i}^s \hspace{-0.2em}=\hspace{-0.2em} F_{g}(s_{i^*}, f_{g_i})$. To achieve hierarchical Gaussian grouping, we employ HDBSCAN to cluster the hierarchical feature Gaussians directly, obtaining Gaussian groups denoted as $\{G_{i}\}$ based on the hierarchical instance features. The central feature of each cluster is denoted as $\{T_i\}$. To select the Gaussian group for complete object localization, our method employs alpha blending (Eqn.~\ref{eq:rendering_f3dgs}) to render the hierarchical instance features of the reference view. We select the instance feature of the pixel with the highest relevancy score $r_i$ as the reference feature $\hat{T_i}$. We determine the final Gaussian group $G_{i}^*$ corresponding to the clustering center features $\{T_i\}$ that are most relevant to the reference feature $\hat{T_i}$:
\begin{equation}
	G_{i^*} = \{ G_i \mid T_i = \arg\max_{T_j} \frac{\hat{T_i} \cdot T_j}{\|\hat{T_i}\| \|T_j\|} \}
	\label{gssim_f3dgs}
\end{equation}
For each novel view, our proposed ReasonGrounder can render the Gaussian group $G_{s_{i^*}}$ to achieve amodal perception, localizing complete objects with occlusions.

\section{Experiments}

% \subsection{Settings}
\noindent\textbf{Datasets.} 
We use the LERF dataset~\cite{kerr2023lerf}, the 3D-OVS dataset~\cite{liu2023weakly}, and our proposed ReasoningGD dataset.

\noindent\textit{(1) LERF dataset.} The LERF dataset consists of 13 scenes, including in-the-wild scenarios and posed long-tail scenes. The scenes were captured using the Polycam iPhone app, employing on-board SLAM for camera pose estimation, and feature images with a resolution of 994$\times$738.

\noindent\textit{(2) 3D-OVS dataset.} The 3D-OVS dataset comprises a collection of long-tail objects captured in diverse poses and backgrounds. This dataset is mainly developed for open-set 3D semantic segmentation with a full list of categories.

\noindent\textit{(3) ReasoningGD dataset.} This paper introduces a novel dataset, ReasoningGD, which includes over 10K scenes of varying complexity and more than 263 types of common objects, with around 2 million annotations. Each scene is generated using the Blenderproc~\cite{denninger2019blenderproc} toolkit and contains 100 RGB-D images, annotated with object labels, camera poses, and 2D modal and amodal binary masks (for both visible and occluded parts). The dataset features multiple object instances with varying levels of occlusion, making it ideal for evaluating the ability in open-vocabulary 3D reasoning, grounding, and amodal perception of novel views.
 
% \end{itemize}

\noindent \textbf{Evaluation Metrics.} The performance of ReasonGrounder is evaluated using two main metrics: Localization Accuracy~\cite{kerr2023lerf} and Intersection over Union (IoU). Localization is deemed successful if the pixel with the highest relevance falls within the labeled bounding box in LERF. The IoU score is calculated across the LERF, 3D-OVS, and ReasoningGD datasets to assess the accuracy of localization for each language category in open-vocabulary 3D localization.

\noindent\textbf{Implementation Details.} To extract language features from each image, we use the OpenCLIP ViT-B/16 model. For SAM, we employ the ViT-H model to segment 2D masks. Each scene is first trained using 3D Gaussian Splatting for 30,000 iterations with default parameters from~\cite{kerbl20233d}. We then train the hierarchical feature Gaussian field by fixing all other parameters of the 3D Gaussians. During this stage, only the latent features, language mapper, and instance mapper are learnable, with the latent features trained for 10,000 iterations. We use LLaVA 1.5 as the LVLM in our ReasonGrounder. All models are trained on an NVIDIA RTX-3090 GPU and a 14 vCPU Intel(R) Xeon(R) Gold 6330 CPU @ 2.00GHz.

\begin{figure*}[t]
	\centering
	\includegraphics[width=0.9\linewidth]{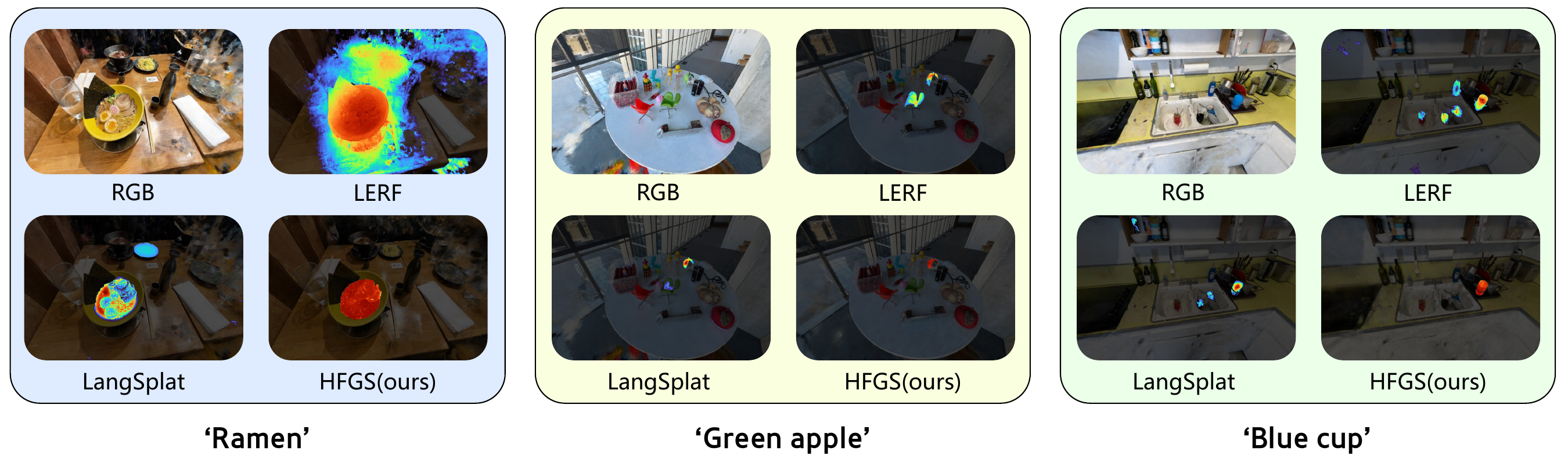}
    \vspace{-0.1in}
	\caption{\textbf{Qualitative comparisons of open-vocabulary 3D visual grounding.} Our ReasonGrounder demonstrates superior accuracy in open-vocabulary 3D localization compared to other state-of-the-art methods.
    }
	\label{fig:explicit_localization}
	  \vspace{-0.15in}
\end{figure*}

\subsection{Evaluation on Open-set 3D Visual Grounding}
Since no open-vocabulary 3D visual grounding and reasoning methods exist for comparison, we evaluate our approach against current open-vocabulary 3D grounding methods.

\begin{table}[t] \small
	\setlength{\tabcolsep}{1pt} % Adjust column spacing
	\centering
	\small % Reduce font size
	\begin{tabular}{lccccc}
		\toprule
		\textbf{Method} & \textit{ramen} & \textit{figurines} & \textit{teatime} & \textit{kitchen} & \textbf{overall}  \\
		\midrule
		LSeg~\cite{li2022language} & 14.1 & 8.9 & 33.9 & 27.3 & 21.1\\
		LERF~\cite{kerr2023lerf} & \cellcolor{softyellow}62.0 & \cellcolor{softyellow}75.0 & \cellcolor{softyellow}84.8 & \cellcolor{softyellow}72.7 & \cellcolor{softyellow}73.6\\
		LangSplat~\cite{qin2024langsplat} & \cellcolor{softorange}73.2 & \cellcolor{softorange}80.4 & \cellcolor{softorange}88.1 & \cellcolor{softorange}95.5 & \cellcolor{softorange}84.3\\
		\textbf{ReasonGrounder(Ours)} & \cellcolor{softred}\textbf{78.5} & \cellcolor{softred}\textbf{82.4} & \cellcolor{softred}\textbf{89.7} & \cellcolor{softred}\textbf{96.2} & \cellcolor{softred}\textbf{86.7} \\
		\bottomrule
	\end{tabular}
      \vspace{-0.1in}
	\caption{\textbf{Localization Accuracy (\%) on the LERF dataset for open-vocabulary 3D visual grounding.}  Our ReasonGrounder employs the same explicit queries as previous  approaches.}
	\label{table:lerf_loca}
  \vspace{-0.15in}
\end{table}

\begin{table}[t] \small
	\setlength{\tabcolsep}{1pt} % 调整列间距
	\centering
	\small % 缩小字体
	\begin{tabular}{lccccc}
		\toprule
		\textbf{Method} & \textit{ramen} & \textit{figurines} & \textit{teatime} & \textit{kitchen} & \textbf{overall}  \\
		\midrule
		LSeg~\cite{li2022language} & 7.0 & 7.6 & 21.7 &29.9 & 16.6\\
		LERF~\cite{kerr2023lerf} & \cellcolor{softyellow}28.2 & \cellcolor{softyellow}38.6 & \cellcolor{softyellow}45.0 & \cellcolor{softyellow}37.9 & \cellcolor{softyellow}37.4\\
		LangSplat~\cite{qin2024langsplat} & \cellcolor{softorange}51.2 & \cellcolor{softorange}44.7 & \cellcolor{softorange}65.1 & \cellcolor{softorange}44.5 & \cellcolor{softorange}51.4\\
		\textbf{ReasonGrounder(Ours)} & \cellcolor{softred}\textbf{53.4} & \cellcolor{softred}\textbf{49.6} & \cellcolor{softred}\textbf{68.2} & \cellcolor{softred}\textbf{49.3} & \cellcolor{softred}\textbf{55.1} \\
		\bottomrule
	\end{tabular}
       \vspace{-0.1in}
	\caption{\textbf{Mean IoU (\%) on  LERF  for open-vocabulary 3D visual grounding.} Our ReasonGrounder employs the same explicit queries as previous state-of-the-art approaches. \label{table:lerf_seg} }
    \vspace{-6mm}
	
\end{table}

\begin{table}[t]
	\renewcommand\tabcolsep{3pt}
	\centering
	\begin{tabular}{lcccccc}
		\toprule
		Method & \textit{bed}   & \textit{bench} & \textit{room}  & \textit{sofa}  & \textit{lawn}  & \textbf{overall}   \\
		\midrule
		LSeg~\cite{li2022language} & 56.0 & 6.0 & 19.2 & 4.5 & 17.5 & 20.6 \\
		ODISE~\cite{xu2023open} & 52.6 & 24.1 & 52.5 & 48.3 & 39.8 & 43.5 \\
		OV-Seg~\cite{liang2023open} & 79.8 & 88.9 & 71.4 & 66.1 & 81.2 & 77.5 \\
		\midrule
		FFD~\cite{kobayashi2022decomposing} & 56.6 & 6.1 & 25.1 & 3.7 & 42.9 & 26.9 \\
		LERF~\cite{kerr2023lerf} & 73.5 & 53.2 & 46.6 & 27 & 73.7 & 54.8 \\
		3D-OVS~\cite{liu2023weakly} & \cellcolor{softyellow}89.5 & \cellcolor{softyellow}89.3 & \cellcolor{softyellow}92.8 & \cellcolor{softyellow}74 & \cellcolor{softyellow}88.2 & \cellcolor{softyellow}86.8 \\
		LangSplat~\cite{qin2024langsplat} & \cellcolor{softorange}92.5 & \cellcolor{softorange}94.2 & \cellcolor{softorange}94.1 & \cellcolor{softorange}90.0 & \cellcolor{softorange}96.1 & \cellcolor{softorange}93.4 \\ 
		\textbf{ReasonGrounder} & \cellcolor{softred}\textbf{93.3} & \cellcolor{softred}\textbf{96.6} & \cellcolor{softred}\textbf{94.5} & \cellcolor{softred}\textbf{91.7} & \cellcolor{softred}\textbf{97.3} & \cellcolor{softred}\textbf{94.7} \\ 		
		\bottomrule
	\end{tabular}
       \vspace{-0.1in}
	\caption{\textbf{Mean IoU scores (\%) on  3D-OVS dataset for open-vocabulary 3D visual grounding.} The first three methods target 2D visual grounding, whereas the remaining methods, including our ReasonGrounder, focus on 3D visual grounding.}
	\label{table:3dovs}
	   \vspace{-0.15in}
\end{table}

\noindent \textbf{Quantitative Results.} We first compare our proposed method with other approaches on the LERF dataset. Table~\ref{table:lerf_loca} shows that our method achieves an overall localization accuracy of 86.7\%, outperforming existing methods. Table~\ref{table:lerf_seg} presents the IoU results for open-vocabulary 3D visual grounding, where our method improves from 51.4\% to 55.1\% compared to LangSplat, demonstrating the superiority of ReasonGrounder.
To further validate the performance of ReasonGrounder, we compare it with other 2D and 3D state-of-the-art methods on the 3D-OVS dataset (Table~\ref{table:3dovs}). Our results show that ReasonGrounder outperforms 2D-based methods like ODISE~\cite{xu2023open} and OV-Seg~\cite{liang2023open}, and significantly surpasses 3D-based methods, including LERF~\cite{kerr2023lerf}, 3D-OVS~\cite{liu2023weakly}, and LangSplat~\cite{qin2024langsplat}. The results highlight that our approach effectively creates precise scale-hierarchical feature Gaussian fields, enabling accurate open-vocabulary 3D visual grounding.

\noindent \textbf{Qualitative Results.} We show the superior quality of open-vocabulary 3D visual grounding in ReasonGrounder compared to other methods in challenging and realistic 3D scenes. Figure~\ref{fig:explicit_localization} illustrates the visual results of this comparison. We note that the activation areas generated by other methods are often more scattered, while our approach yields more focused regions. Furthermore, our activation regions align more closely with the ground truth shape.

\subsection{Evaluation on 3D Reasoning}
Compared to existing methods, ReasonGrounder enables open-vocabulary 3D reasoning, enhancing implicit instruction comprehension and amodal perception of novel views. This section highlights its superior capabilities in 3D localization and perception under novel views.

\begin{figure*}
	\centering
	\includegraphics[width=0.9\linewidth]{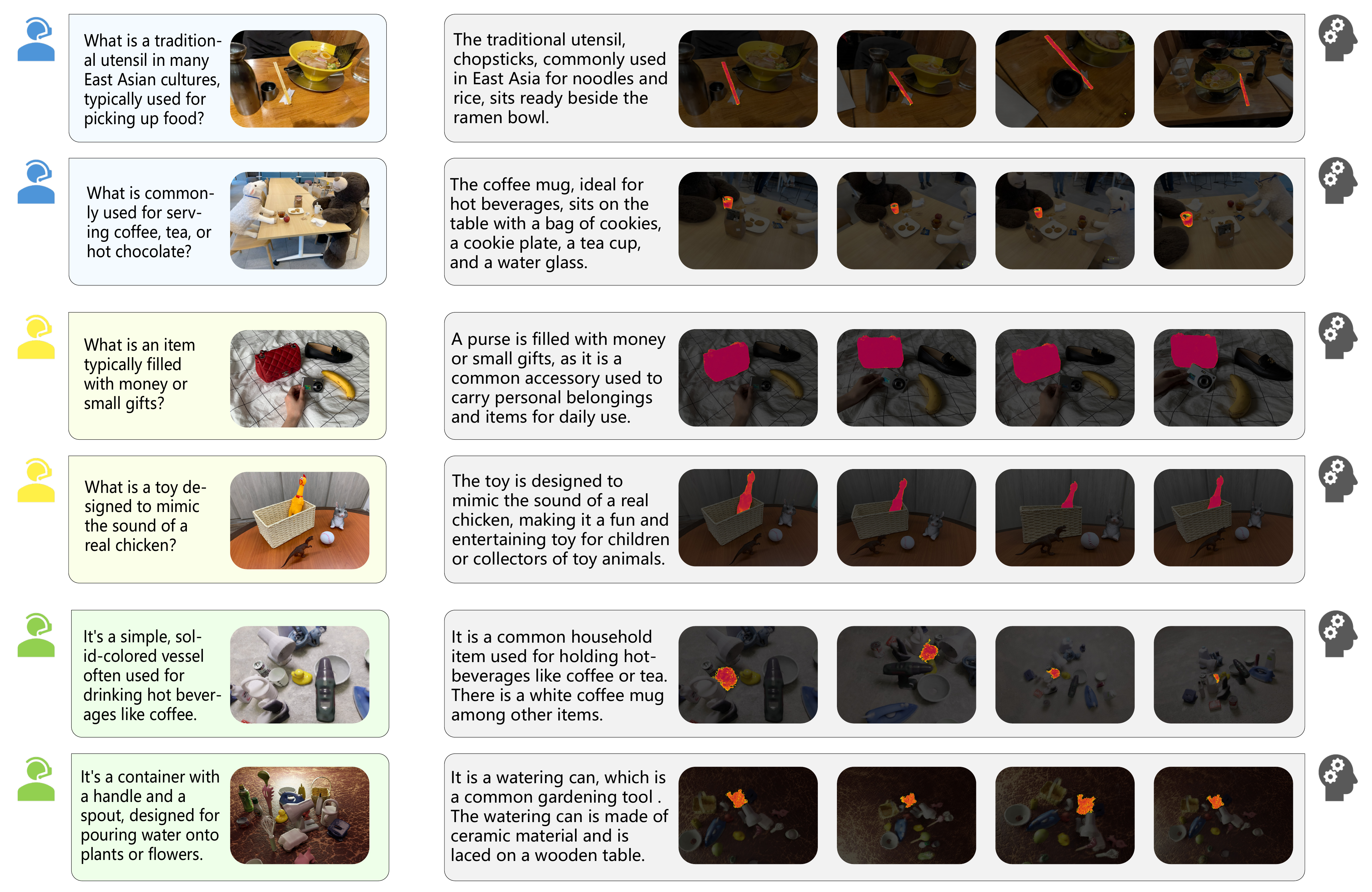}
        	\vspace{-0.1in}
	\caption{\textbf{Qualitative results of 3D localization with implicit instructions on the LERF, 3D-OVS, and ReasoningGD datasets.} These results demonstrate that our \textbf{ReasonGrounder} can accurately interpret implicit instructions and identify the target object.\label{fig:implicit_localization}}
	\vspace{-0.2in}
\end{figure*}

\begin{figure*}
	\centering
	\includegraphics[width=\linewidth]{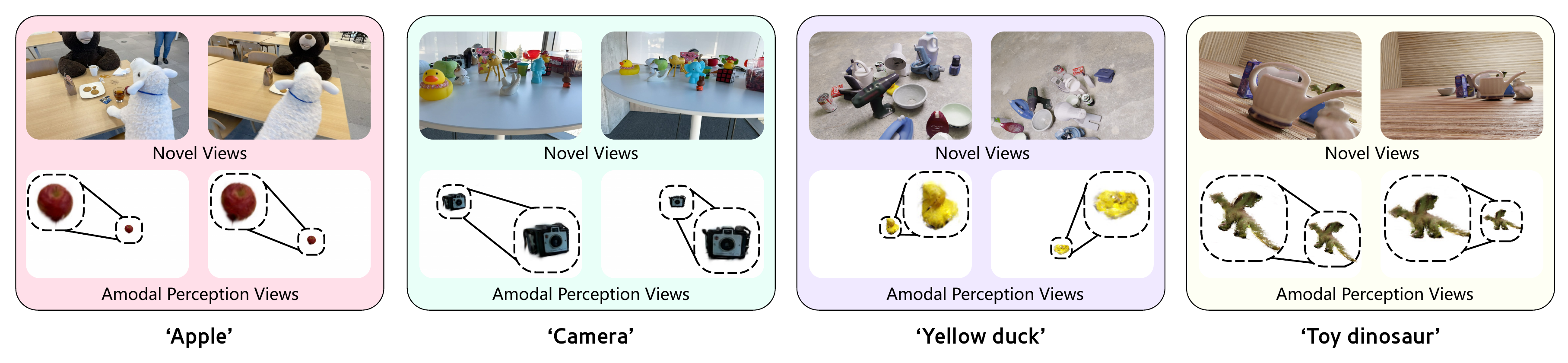}
    	\vspace{-0.25in}
	\caption{\textbf{Qualitative results of amodal perception of novel views on the LERF and proposed ReasoningGD datasets.} Our \textbf{ReasonGrounder} achieves accurate 3D localization, even when the object is partially visible or fully occluded in novel views.\label{fig:amodal_localization} }

		\vspace{-0.15in}
\end{figure*}

\begin{table}[t] \small
\renewcommand{\tabcolsep}{1pt} % 缩小列间距
\centering
\begin{tabular}{lcccccc}
\toprule
{\textbf{Dataset}}  & \multicolumn{6}{c}{\textbf{Scene}} \\ 
\midrule
\multirow{2}{*}{\shortstack[l]{LERF}}        & \textit{ramen} & \textit{figurines} & \textit{teatime} & \textit{kitchen} & \textit{-} & \textbf{overall} \\ 
                                     & 53.8  & 49.5      & 67.9    & 49.6    & - & 55.2    \\
\midrule
\multirow{2}{*}{\shortstack[l]{3D-OVS}}      & \textit{bed}   & \textit{bench}     & \textit{room}    & \textit{sofa}    & \textit{lawn} & \textbf{overall}  \\ 
                                     & 93.1  & 94.8      & 92.9    & 91.4    & 96.8 & 93.8     \\
\midrule
\multirow{2}{*}{\shortstack[l]{ReasoningGD}} & \textit{001}   & \textit{002}       & \textit{003}     & \textit{004}     & \textit{005}  & \textbf{overall}  \\ 
                                     & 91.8  & 87.4      & 93.6    & 91.3    & 92.8 & 91.4     \\
\bottomrule
\end{tabular}
    \vspace{-0.1in}

\caption{\textbf{Quantitative results of mean IoU (\%) across various scenes in the LERF, 3D-OVS, and ReasoningGD datasets, including both scene-specific scores and overall performance.} This highlights the performance of our ReasonGrounder in open-vocabulary 3D visual grounding with implicit queries.}
\label{table:im_seg}
\vspace{-6mm}
\end{table}

\noindent \textbf{3D Localization with Implicit Instructions.} ReasonGrounder leverages LVLM for implicit queries requiring complex reasoning, selecting Gaussian groups for 3D localization. Instead of explicit labels like ``Coffee mug,” we use implicit queries from prior annotations~\cite{qin2024langsplat}, e.g., ``Which object can hold coffee?” Table~\ref{table:im_seg} quantifies the method’s effectiveness across five ReasoningGD scenes (001-005).
Figure~\ref{fig:implicit_localization} demonstrates the effectiveness of 3D visual grounding with implicit, complex text queries in realistic 3D scenes. The qualitative results confirm that ReasonGrounder excels in localizing target objects, benefiting from the 3D consistency. Additionally, ReasonGrounder provides explanatory answers, showcasing its strength in implicit instruction reasoning, 3D understanding, and conversation. To test robustness, we selected five challenging scenes with small proportions, including multi-hierarchical structures and similar objects, along with ten text queries per scene from the LERF and ReasoningGD datasets. As shown in Table~\ref{table:challenge}, our method outperforms in all edge cases.

\noindent \textbf{Amodal Perception of Novel Views.} Existing open-vocabulary 3D visual grounding methods struggle with localizing complete objects in novel views with occlusion, limiting their real-world applicability. In contrast, ReasonGrounder uses hierarchical Gaussian grouping to effectively tackle this issue. The ReasoningGD dataset introduced here includes scenes with amodal binary mask annotations, accurately representing the full shape of occluded objects from different views. Quantitative results from five typical scenes, shown in Table~\ref{table:lerf_amodal_im_seg}, demonstrate that ReasonGrounder excels in amodal perception, successfully localizing target objects even when occluded in novel views. For qualitative experiments, we present ReasonGrounder's evaluation on the LERF and ReasoningGD datasets in Figure~\ref{fig:amodal_localization}. For each query, the left column shows partial occlusion, and the right column shows full occlusion. These results demonstrate that ReasonGrounder successfully achieves amodal perception, accurately localizing complete objects regardless of the occlusion level.
\begin{table}[t]
    \small
	\vspace{-7pt}
	\renewcommand\tabcolsep{3pt}
	\centering
	\begin{tabular}{lcccccc}
		\toprule
		Method & \textit{1}   & \textit{2} & \textit{3}  & \textit{4}  & \textit{5}  & \textbf{overall}   \\
		\midrule
		LSeg & 16.0 & 5.0 & 15.2 & 3.4 & 13.2 & 10.6 \\
		LERF & 62.4 & 51.7 & 47.4 & 23.2 & 55.7 & 48.1 \\
		3D-OVS & \cellcolor{softyellow}66.5 & \cellcolor{softyellow}63.3 & \cellcolor{softyellow}59.4 & \cellcolor{softorange}32.2 & \cellcolor{softorange}60.4 & \cellcolor{softyellow}56.4 \\
		LangSplat & \cellcolor{softorange}70.3 & \cellcolor{softorange}71.2 & \cellcolor{softorange}63.4 & \cellcolor{softyellow}30.4 & \cellcolor{softyellow}59.4 & \cellcolor{softorange}58.9 \\ 
		\textbf{ReasonGrounder} & \cellcolor{softred}\textbf{82.4} & \cellcolor{softred}\textbf{81.8} & \cellcolor{softred}\textbf{85.5} & \cellcolor{softred}\textbf{54.6} & \cellcolor{softred}\textbf{88.3} & \cellcolor{softred}\textbf{78.5} \\ 
		\bottomrule
	\end{tabular}
	\vspace{-9pt}
\caption{\textbf{Quantitative results of mean IoU (\%) across challenge scenes in the LERF and ReasoningGD datasets.} This highlights the robustness of our ReasonGrounder in complex situations.}
\label{table:challenge}
	\vspace{-4mm}
\end{table}
\begin{table} \small
	\setlength{\tabcolsep}{4pt} % Adjust column spacing
	\centering
	\small % Reduce font size
	\begin{tabular}{lcccccc}
		\toprule
		Method & 001 & 002 & 003 & 004 & 005 & \textbf{overall}  \\
		\midrule
		ReasonGrounder & 90.7 & 88.2 & 91.5 & 89.4 & 92.3 & 90.4 \\
		\bottomrule
	\end{tabular}
    \vspace{-0.1in}
	\caption{\textbf{Quantitative results of mean IoU scores (\%) for amodal perception in novel views using the ReasoningGD dataset.} The ReasoningGD dataset provides complete masks of occluded objects as ground truth, enabling quantitative evaluation.\label{table:lerf_amodal_im_seg} }
	\vspace{-6mm}
\end{table}

\subsection{Ablation Study}
Ablations on the figurine scene (Table~\ref{table:ablation_combined}) were tested on an NVIDIA H100 GPU. Without our components, using NeRF for open-vocabulary 3D visual grounding (\textbf{O-3DVG}) achieves 47.2\% IoU with 0.92s per view. Replacing NeRF with 3D Gaussian Splatting (3DGS) boosts IoU by 2.6\% and speeds up rendering to 0.895s.
ReasonGrounder integrates LVLM for implicit instruction reasoning, enabling 3D localization (\textbf{I-3DVG}). Adding scale-hierarchical feature Gaussian grouping (\textbf{SHF}) further enables amodal perception (\textbf{AP}) of novel views. The Ramen scene lacks amodal mask ground truth, so its results are omitted.
Ablations were also conducted on the ReasoningGD dataset, which includes mask annotations for visible and obscured areas. Table~\ref{table:ablation_combined} reports results for scene 001.
\begin{table}[t] \small
	\vspace{-6pt}
	\renewcommand\tabcolsep{4pt} % reduce column spacing
	\centering
	%\small % reduce font size
	\begin{tabular}{ccc|ccc}
		\toprule
		\multicolumn{6}{c}{\textbf{Results on Figurines Scene}} \\
		\midrule
		\multicolumn{3}{c|}{Component} & \multicolumn{3}{c}{Performance} \\
		\midrule
		SHF & LVLM & 3DGS & O-3DVG & I-3DVG & AP \\
		\midrule
		&  &  & 47.2/0.924 & \XSolidBrush & \XSolidBrush \\
		&  & \CheckmarkBold & 49.8/0.025 & \XSolidBrush & \XSolidBrush \\
		& \CheckmarkBold & \CheckmarkBold & 49.3/0.032 & 48.9/0.061 & \XSolidBrush \\
		\CheckmarkBold & \CheckmarkBold & \CheckmarkBold & {53.4/0.053} & {53.8/0.082} & \CheckmarkBold \\
		\midrule
		\multicolumn{6}{c}{\textbf{Results on 001 Scene}} \\
		\midrule
		\multicolumn{3}{c|}{Component} & \multicolumn{3}{c}{Performance} \\
		\midrule
		SHF & LVLM & 3DGS & O-3DVG & I-3DVG & AP \\
		\midrule
		&  &  & 85.4/0.891 & \XSolidBrush & \XSolidBrush \\
		&  & \CheckmarkBold & 87.8/0.026 & \XSolidBrush & \XSolidBrush \\
		& \CheckmarkBold & \CheckmarkBold & 87.7/0.028 & 86.9/0.055 & \XSolidBrush \\
		\CheckmarkBold & \CheckmarkBold & \CheckmarkBold & {91.6/0.061} & {91.8/0.085} & {90.7/0.091} \\
		\bottomrule
	\end{tabular}	
    \vspace{-0.1in}
    \caption{\textbf{Ablation studies.} The results are presented for two different scenes: the Figurines scene from the LERF dataset and the 001 scene from the proposed ReasoningGD dataset.}
	\label{table:ablation_combined}
	\vspace{-19pt}
\end{table}
Quantitative ablation results validate ReasonGrounder’s amodal perception. By exploring open-vocabulary 3D visual grounding and reasoning, ReasonGrounder enables accurate localization with implicit instructions and amodal perception, significantly enhancing its applicability in real-world environments.

\section{Conclusion}
% We propose ReasonGrounder, a novel LVLM-guided framework that leverages hierarchical 3D feature Gaussian fields for adaptive Gaussian grouping with physical scale, enabling effective open-vocabulary 3D visual grounding and reasoning. By using a scale-hierarchical feature Gaussian field, it enhances 3D visual grounding and Gaussian grouping. Leveraging LVLM, ReasonGrounder interprets complex user intent and localizes occluded objects through selecting the feature Gaussian group. We also present the ReasoningGD dataset with over 10K scenes and 2 million annotations. Experimental results demonstrate ReasonGrounder's superiority in open-vocabulary 3D visual grounding, implicit instruction understanding, and amodal perception from novel views.

We propose a novel method ReasonGrounder. By leveraging LVLM, ReasonGrounder interprets complex intent and localizes occluded objects through feature Gaussian selection. We also introduce the ReasoningGD dataset with over 10K scenes and 2 million annotations. Experiments show ReasonGrounder's superiority in open-vocabulary grounding, implicit instruction understanding, and amodal perception from novel views.

\section{Acknowledgments}
This work was supported by the Science and Technology Commission of Shanghai Municipality (No. 24511103100) and in part by NSFC Project (62176061), and Shanghai Technology Development and Entrepreneurship Platform for Neuromorphic and AI SoC.	

{
    \small
    \bibliographystyle{ieeenat_fullname}
    \bibliography{main}
}


\clearpage
\setcounter{page}{1}
\maketitlesupplementary

\setcounter{section}{0}
\section{DATASETS}
This paper selects the LERF dataset, 3D-OVS dataset, and our proposed ReasoningGD dataset for both training and evaluation. The LERF and 3D-OVS datasets are widely used to evaluate the open-vocabulary 3D visual grounding performance of various methods. However, these datasets only provide explicit queries, lacking implicit, descriptive ones. For instance, they utilize a explicit query like “apple” rather than a implicit query such as “red nutrient-rich sweet fruit,” which does not explicitly name the object. To enhance the utility of these two widely used datasets in evaluating our ReasonGrounder’s ability to handle implicit queries, we have added additional annotations for implicit queries. Open-vocabulary 3D visual grounding and reasoning methods must also achieve amodal perception, meaning they should fully identify an object’s structure and shape even when parts of it are occluded from a given angle of view. To address this, we introduce the ReasoningGD dataset, which includes over 10,000 scenes and more than 2 million modal and amodal annotations. These annotations cover both the visible mask area and the occluded portions of objects, enabling more comprehensive evaluation of amodal perception capabilities.

\subsection{Widely used datasets}
\begin{figure}[t]
	\centering
	\includegraphics[width=0.9\linewidth]{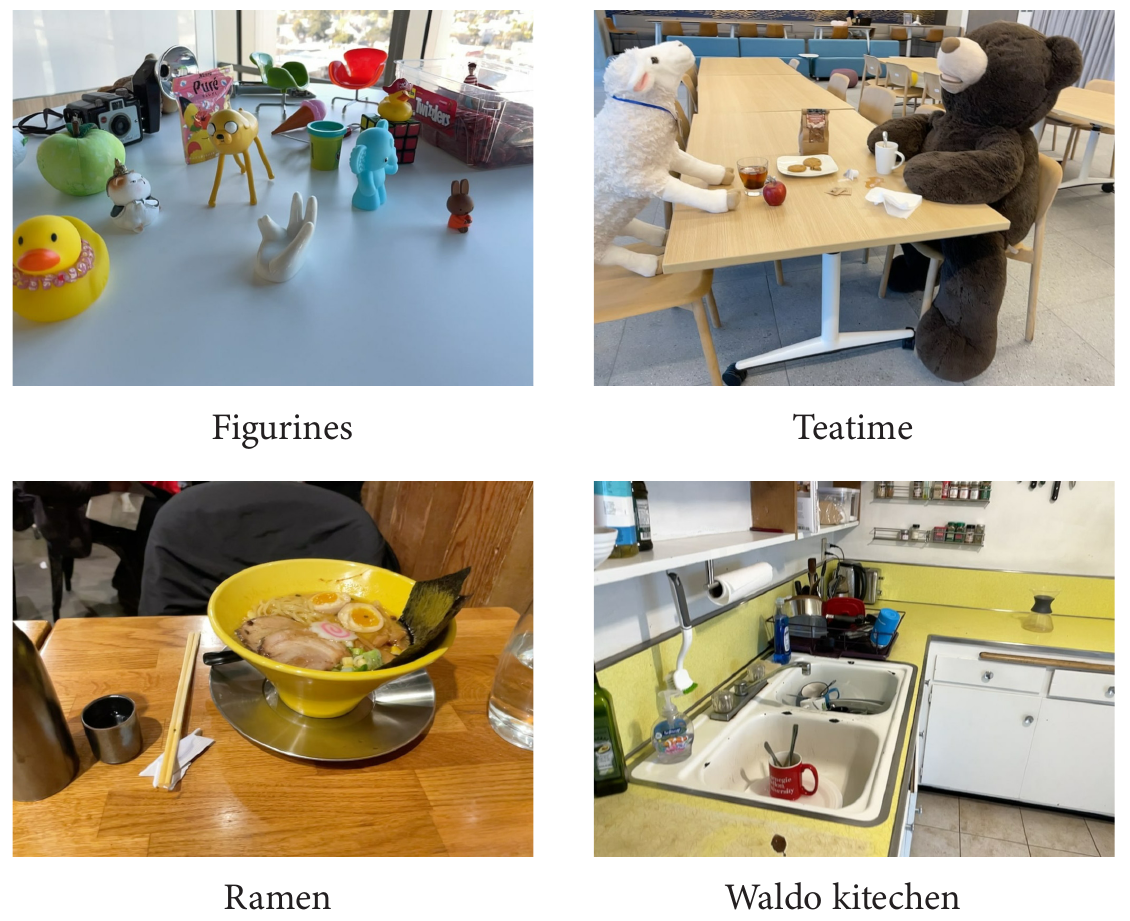}
    \vspace{-0.1in}
	\caption{\textbf{Visualization of LERF dataset.} The LERF dataset comprises various scenes, encompassing both in-the-wild scenarios and posed
long-tail scenes. 
    }
	\label{fig:supple_lerf}
	  \vspace{-0.15in}
\end{figure}

\noindent\textbf{LERF dataset.} As shown in Figure \ref{fig:supple_lerf}, LERF dataset comprises various scenes, encompassing both in-the-wild scenarios and posed
long-tail scenes. These scenes are captured using the Polycam iPhone app and the featured images have a resolution
of 994×738. Thanks to LangSplat’s annotations of the four scenarios in this dataset, we can comprehensively evaluate the open-vocabulary 3D visual grounding capabilities.
\begin{itemize}
    \item \textbf{(1) \textit{Figurines}}: The scene includes various toys and small objects arranged on a round table, such as a rubber duck, blue elephant figurine, Rubik’s Cube, and other colorful items. A container with snack packets is also present. 
    \item \textbf{(2) \textit{Teatime}}: The scene features two large plush toys, a bear and a sheep, seated at a wooden table as if hosting a tea party. The table is set with items like tea, an apple, cookies, and a mug.
    \item \textbf{(3) \textit{Ramen}}: The scene shows a bowl of ramen placed on a wooden table. The vibrant yellow bowl contains slices of pork, half-boiled eggs, narutomaki (fish cake), seaweed, and noodles. Chopsticks and a small cup, likely for soy sauce or sake, are set beside it.
    \item \textbf{(4) \textit{Waldo kitchen}}:The scene depicts a domestic kitchen with a vintage aesthetic, featuring white cabinets with metallic handles and a bright yellow countertop. 
\end{itemize}
To further assess the ability to locate objects based on implicit queries, we have added additional implicit query annotations for these four scenarios. 
Each scene contains ten implicit queries for five objects, totaling 200 implicit queries across all four scenes.
These annotations do not explicitly mention the name of the object to be queried but instead provide descriptive cues related to its characteristics. Below are examples of these scene annotations:
\begin{table}[h]
    \centering
    \renewcommand{\arraystretch}{1.2} % 调整行距
    \setlength{\tabcolsep}{0.5pt} % 调整列间距
    \begin{tabular}{@{}>{\raggedright\arraybackslash}m{4cm} >{\raggedright\arraybackslash}m{4cm}@{}}
        \toprule
        \textbf{Scene} & \textbf{Implicit Query} \\
        \midrule
        \textit{Old camera} in \textbf{Figurines} & \textit{It is a vintage device used for capturing photographs on film.} \\
        \textit{Coffee mug} in \textbf{Teatime} & \textit{It is a sturdy vessel designed for holding hot beverages.} \\
        \textit{Chopsticks} in \textbf{Ramen} & \textit{They are a pair of slender tools used for picking up food.} \\
        \textit{Toaster} in \textbf{Waldo kitchen} & \textit{This is an appliance designed to brown slices of bread using heat.} \\
        \bottomrule
    \end{tabular}
    \caption{\textbf{Examples of annotated implicit queries in LERF dataset.} These implicit annotations do not directly mention the name of the object to be queried; instead, they offer descriptive hints about the object’s characteristics.}
\end{table}

\begin{figure}[t]
	\centering
	\includegraphics[width=0.9\linewidth]{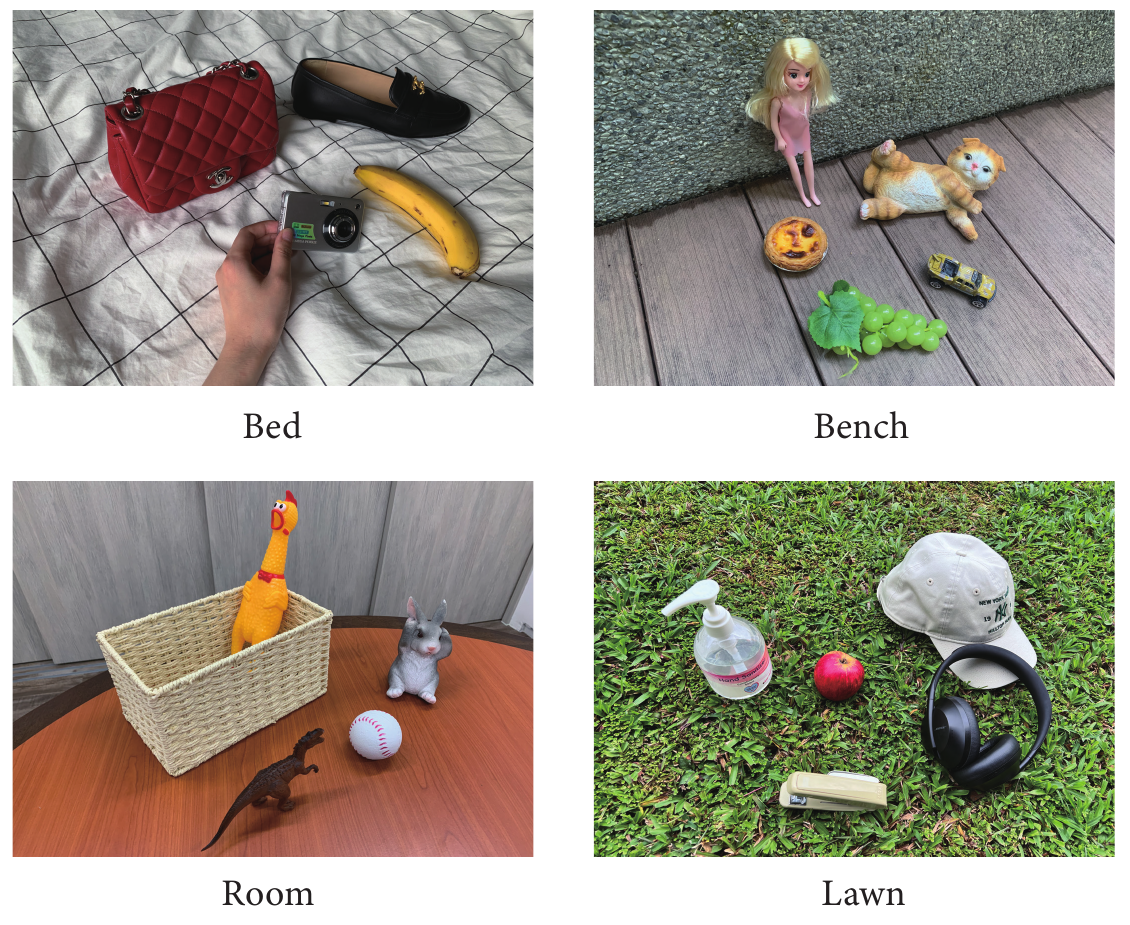}
    \vspace{-0.1in}
	\caption{\textbf{Visualization of 3D-OVS dataset.} Each scene in the 3D-OVS dataset is accompanied by textual descriptions of objects, which guide the segmentation process. 
    }
	\label{fig:supple_3D-OVS}
	  \vspace{-0.15in}
\end{figure}

\noindent\textbf{3D-OVS dataset.} Each scene in the 3D-OVS dataset is accompanied by textual descriptions of objects, which guide the segmentation process. These descriptions are crucial for helping the model identify and segment objects in a scene based on their open-vocabulary labels. As a result, this widely used dataset is well-suited for evaluating open-vocabulary 3D visual grounding methods. As shown in Figure~\ref{fig:supple_3D-OVS}, this paper primarily conducts experiments based on five scenarios.
\begin{itemize}
\item \textbf{(1) \textit{Bed}}: The scene shows a selection of personal items arranged on a checkered fabric surface. Visible items include a red quilted handbag, a black loafer, a banana, and a digital camera held by a hand. The setup suggests a casual, stylish collection of fashion and everyday essentials.
\item \textbf{(2) \textit{Bench}}: The scene displays a small collection of objects on a wooden surface, such as a blonde doll, a toy cat figurine, a miniature toy car, a bunch of green grapes, and an egg tart. The setting is outdoors, with a textured wall in the background.
\item \textbf{(3) \textit{Room}}: The scene presents a playful arrangement of toys on a wooden table, including a rubber chicken in a wicker basket, a small rabbit figurine, a dinosaur toy, and a baseball.
\item \textbf{(4) \textit{Sofa}}: The scene shows a variety of entertainment-related items on a gray surface. These include a plush toy in a festive costume, a stack of UNO cards, a pink gaming controller, a white Xbox controller, and a robot or mecha model.
\item \textbf{(5) \textit{Lawn}}: The scene depicts a collection of items on a grassy surface, including a bottle of hand sanitizer, a red apple, a white baseball cap, a pair of black headphones, and a stapler.
\end{itemize}
To further assess the ability to localize objects based on implicit queries, we have added additional implicit query annotations for these five scenarios. 
Each scene contains ten implicit queries for four objects, totaling 200 implicit queries across all four scenes:
\begin{table}[t]
    \centering
    \renewcommand{\arraystretch}{1.2} % 调整行距
    \begin{tabular}{@{}>{\raggedright\arraybackslash}m{3cm} >{\raggedright\arraybackslash}m{5cm}@{}}
        \toprule
        \textbf{Scene} & \textbf{Implicit Query} \\
        \midrule
        \textit{Red bag} in \textbf{bed} & \textit{Search for a item typically filled with money and gifts.} \\
        \textit{Orange cat} in \textbf{Bench} & \textit{It has a vibrant coat that ranges from light to dark orange.} \\
        \textit{Base ball} in \textbf{Room} & \textit{It is a round object used in a popular bat-and-ball sport.} \\
        \textit{UNO cards} in \textbf{Sofa} & \textit{It is typically played by two or more players.} \\
        \textit{Stapler} in \textbf{Lawn} & \textit{It is a device used to fasten sheets of paper together.} \\
        \bottomrule
    \end{tabular}
    \caption{\textbf{Examples of annotated implicit queries in 3D-OVS dataset.} These implicit annotations do not directly mention the name of the object to be queried; instead, they offer descriptive hints about the object’s characteristics.}
\end{table}

\begin{figure}[t]
	\centering
	\includegraphics[width=0.9\linewidth]{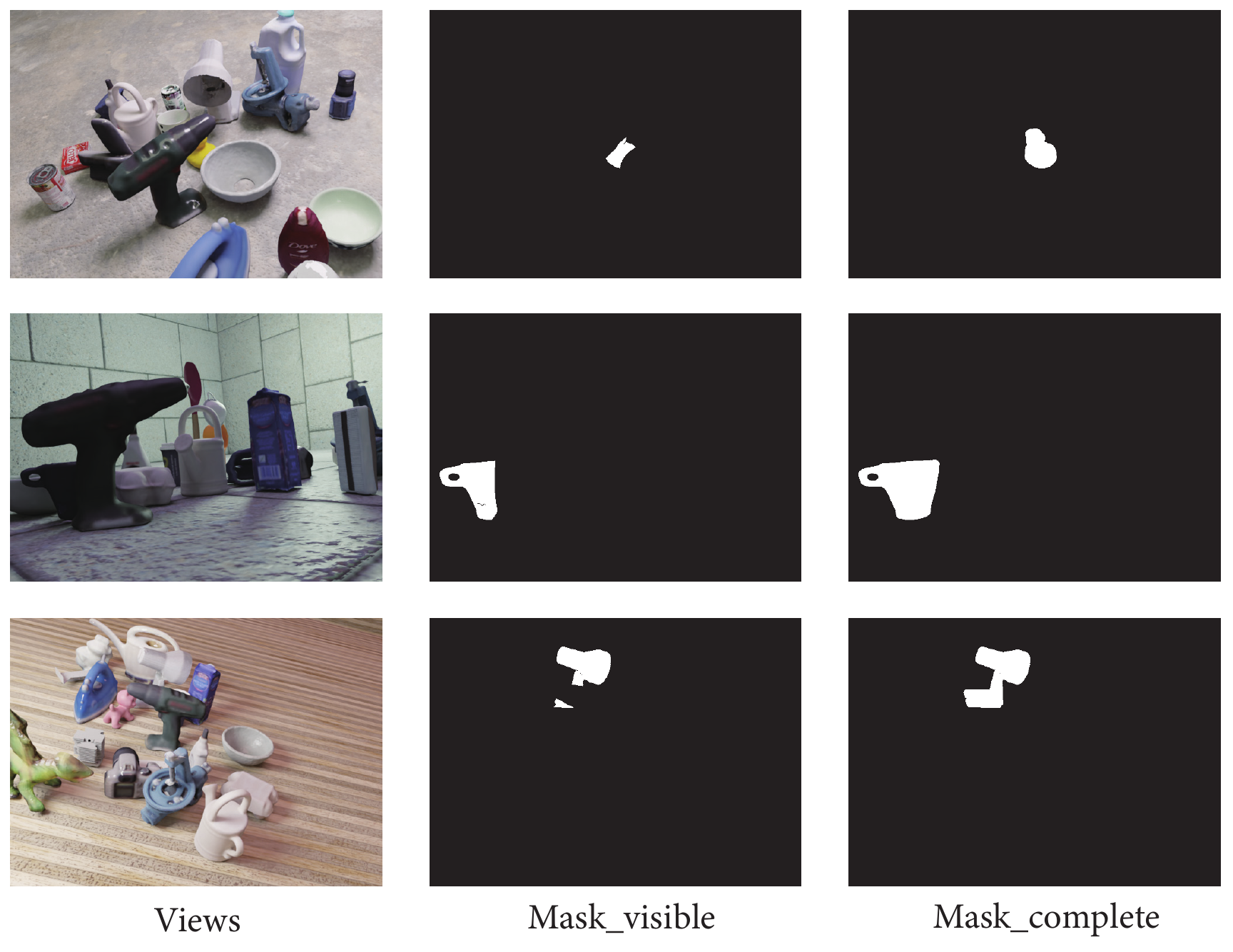}
    \vspace{-0.1in}
	\caption{\textbf{Visualization of proposed ReasoningGD dataset.} ReasoningGD dataset encompasses a diverse range of occlusion scenarios and offers comprehensive, accurate annotations.  
    }
	\label{fig:supple_RGD}
	  \vspace{-0.15in}
\end{figure}
\subsection{Proposed ReasoningGD dataset}
Open-vocabulary 3D visual grounding and reasoning aim to localize objects in a scene based on implicit language descriptions, even when parts of the objects are occluded. However, the existing LERF and 3D-OVS datasets, which are primarily used to evaluate open-vocabulary 3D visual grounding, face limitations in assessing 3D reasoning capabilities. These include challenges in localization based on implicit instructions and identifying the full structural shape of occluded objects. These shortcomings stem from the lack of annotations for implicit instructions and the obscured portions of objects. As a result, these datasets lack the ground truth necessary for effectively evaluating open-vocabulary 3D visual grounding and reasoning tasks.

This paper introduces the ReasoningGD dataset, which encompasses a diverse range of occlusion scenarios and offers comprehensive, accurate annotations. As shown in Figure~\ref{fig:supple_RGD}, these annotations include both the visible and occluded parts of target objects from various perspectives, enabling more robust evaluation of reasoning capabilities in 3D visual grounding. The dataset comprises over 10K scenes, each featuring 10 to 15 objects. Each scene includes 100 viewing angles, with annotations provided for both the visible mask of each object at each angle and the full mask, which includes occluded parts. In total, the dataset contains over 2 million detailed annotations.
The composition of each scene is primarily detailed in the Table \ref{table:dataset}.
\begin{table}[h!]
\centering
\begin{tabular}{@{}m{4cm}<{\centering}m{4cm}<{\centering}@{}}
\toprule
\textbf{Scene Composition} & \textbf{Description} \\ \midrule
Images & Stores the original training views \\ 
Points3d.ply & Includes 3D point cloud data of the scene \\ 
Mask\_visible & The mask annotation of the visible parts of each object from different views \\ 
Mask\_complete & The mask annotation of the full shape and structure of each object from different views \\ 
Transforms\_test.json & Describes transformations for test data \\ 
Transforms\_train.json & Describes transformations for training data \\ 
\bottomrule
\end{tabular}
\caption{\textbf{Composition and description of each scene in the ReasoningGD dataset.} The ReasoningGD dataset consists of diverse scenes designed to evaluate 3D visual grounding and reasoning capabilities under various conditions. Each scene is carefully structured with the key components.}
\label{table:dataset}
\end{table}

\section{IMPLEMENTATION DETAILS}
Given training views with corresponding camera poses, we first construct a standard 3D Gaussian Splatting field. The training parameters used in this process align with those specified in the original paper. Across various scenes, ReasonGrounder employs consistent hyperparameters for uniformity.
Each Gaussian in the field is assigned a 32-dimensional latent feature. This latent feature is subsequently mapped to hierarchical language features and hierarchical instance features. For object segmentation, we utilize the SAM ViT-H model to process the training views, generating object masks for each view.
To supervise the hierarchical language features, we introduce the OpenCLIP ViT-B/16 model, extracting CLIP features of all object masks. The CLIP feature space has a dimensionality of 512. However, due to the computational burden posed by mapping the latent features of all Gaussians to 512 dimensions, we employ Principal Component Analysis (PCA) to reduce the dimensionality of the CLIP features to 64. These compressed 64-dimensional features are then used as supervision signals for the hierarchical language features of each Gaussian.
The compression matrix obtained through PCA is retained, enabling decompression to restore the 64-dimensional hierarchical language features back to their original 512-dimensional space. This ensures efficient computation during training while maintaining fidelity for open-vocabulary 3D visual grounding during rendering.

For implicit queries, we introduce the Large Vision-Language Model (LVLM) to comprehend and reason about the target object. Specifically, we adopt the LLaVA-v1.5-7B model for this purpose. Using the inferred target object, the hierarchical language feature and instance feature are retrieved.
To localize the target object, the hierarchical feature space is clustered using HDBSCAN (Hierarchical Density-Based Spatial Clustering of Applications with Noise). This approach determines the corresponding Gaussian group, effectively identifying the target object, even when occlusion is present in the current view.
HDBSCAN is initialized with fixed parameters to ensure robust clustering:
\begin{itemize}
    \item \textbf{{min\_cluster\_size=10}}: Clusters must contain at least 10 points to be valid.
    \item \textbf{{cluster\_selection\_epsilon=0.01}}: Ensures strict cluster boundaries for precision.
    \item \textbf{{allow\_single\_cluster=False}}: Prevents all points from being grouped into a single large cluster, treating poorly clustered points as noise instead.
\end{itemize}
This configuration enhances the clustering process, enabling accurate localization of objects while maintaining sensitivity to occlusion.
\section{HIERARCHICAL FEATURE}
\begin{figure*}
    \centering
    \includegraphics[width=\linewidth]{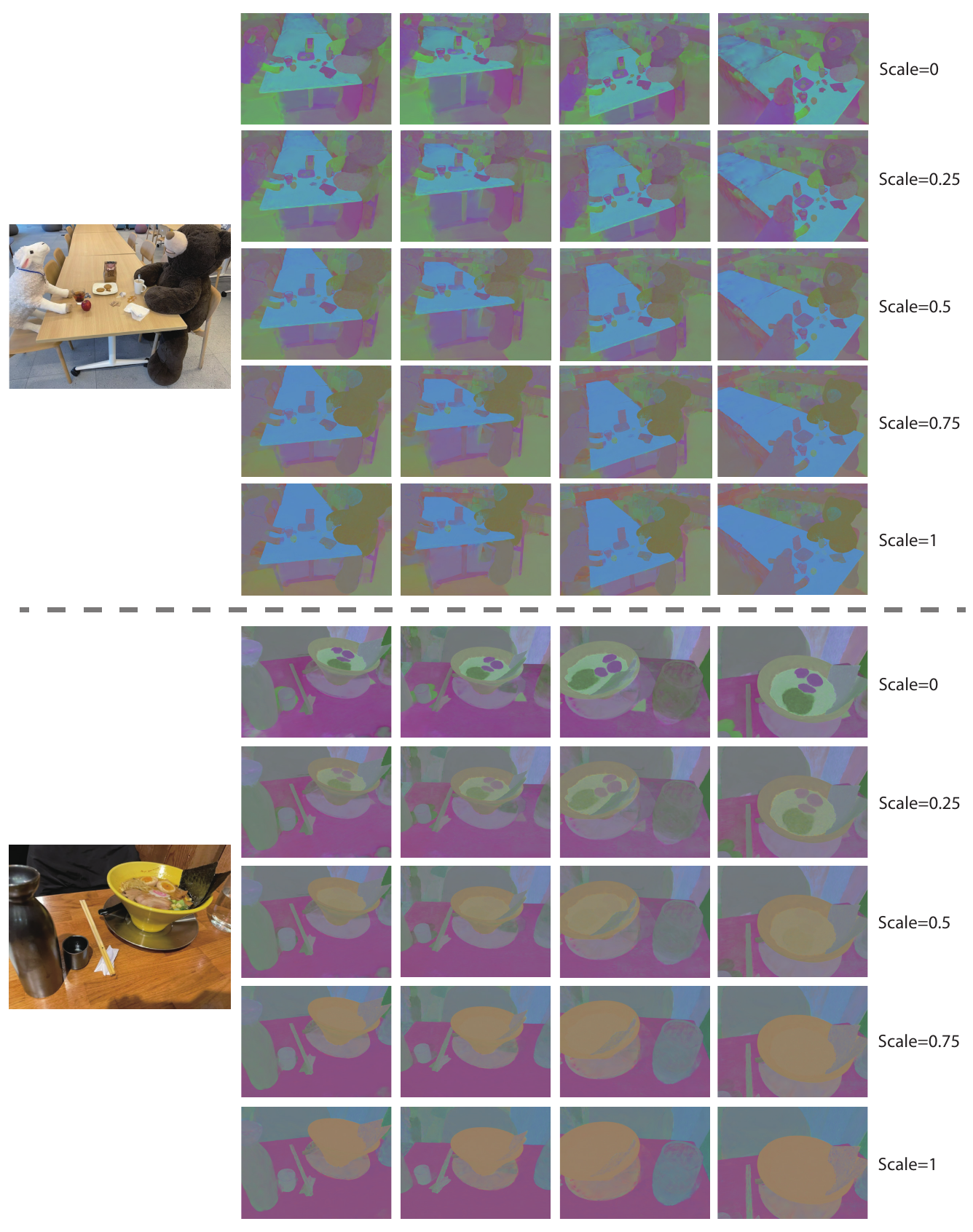}
        \vspace{-0.3in}
	\caption{\textbf{Visualization of hierarchical feature.} The feature space dynamically adjusts its granularity in response to physical scale: finer granularity is employed at smaller scales to capture intricate details, while coarser granularity is utilized at larger scales to emphasize overarching structures. This adaptive mechanism ensures robust and effective representation across a wide range of scales.\label{hlerf} }	
	 \vspace{-0.2in}
\end{figure*}

\begin{figure*}
    \centering
    \includegraphics[width=\linewidth]{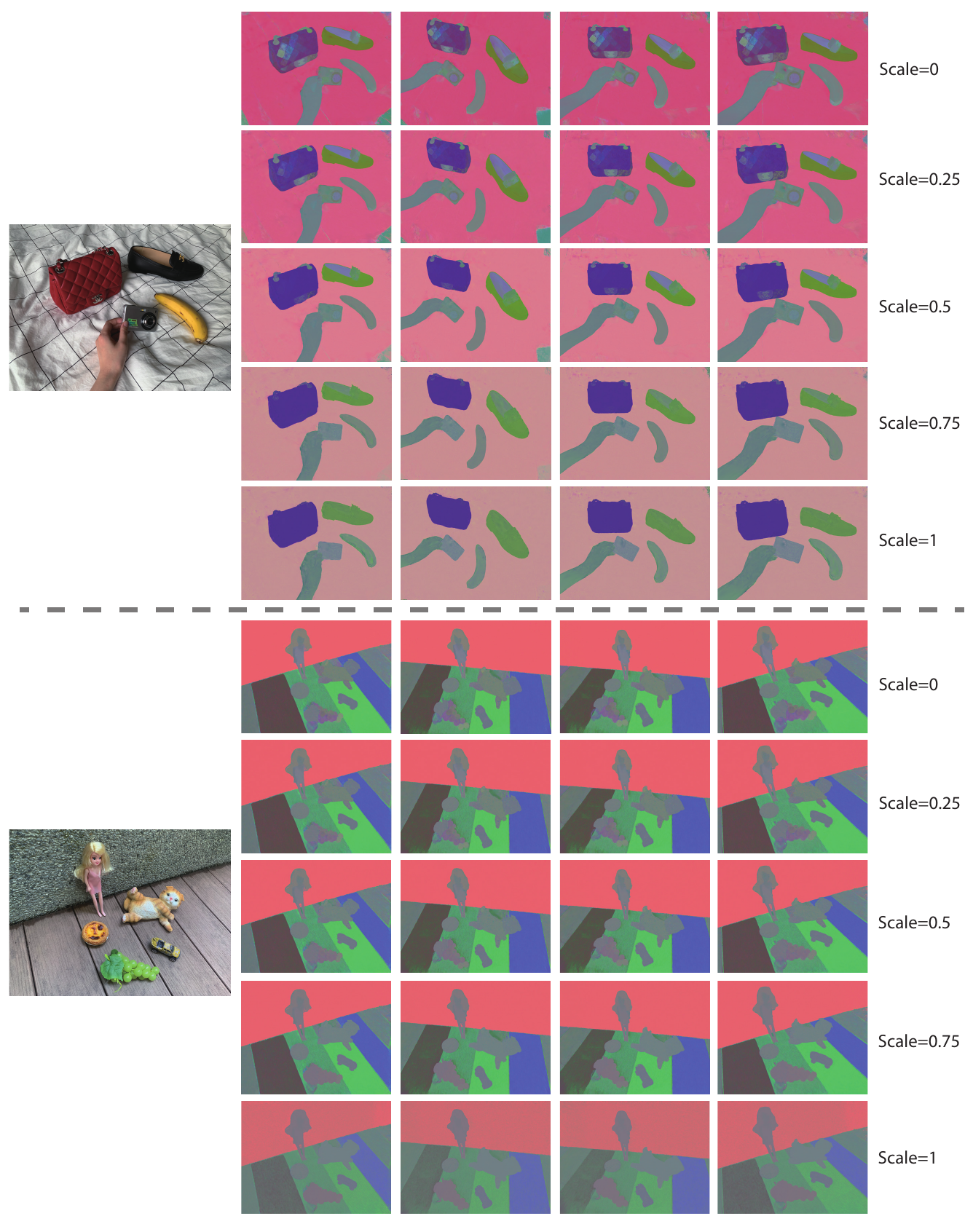}
        \vspace{-0.3in}
\caption{\textbf{Visualization of hierarchical feature.} The feature space dynamically adjusts its granularity in response to physical scale: finer granularity is employed at smaller scales to capture intricate details, while coarser granularity is utilized at larger scales to emphasize overarching structures. This adaptive mechanism ensures robust and effective representation across a wide range of scales.\label{hovs} }	
	 \vspace{-0.2in}
\end{figure*}

\begin{figure*}
    \centering
    \includegraphics[width=\linewidth]{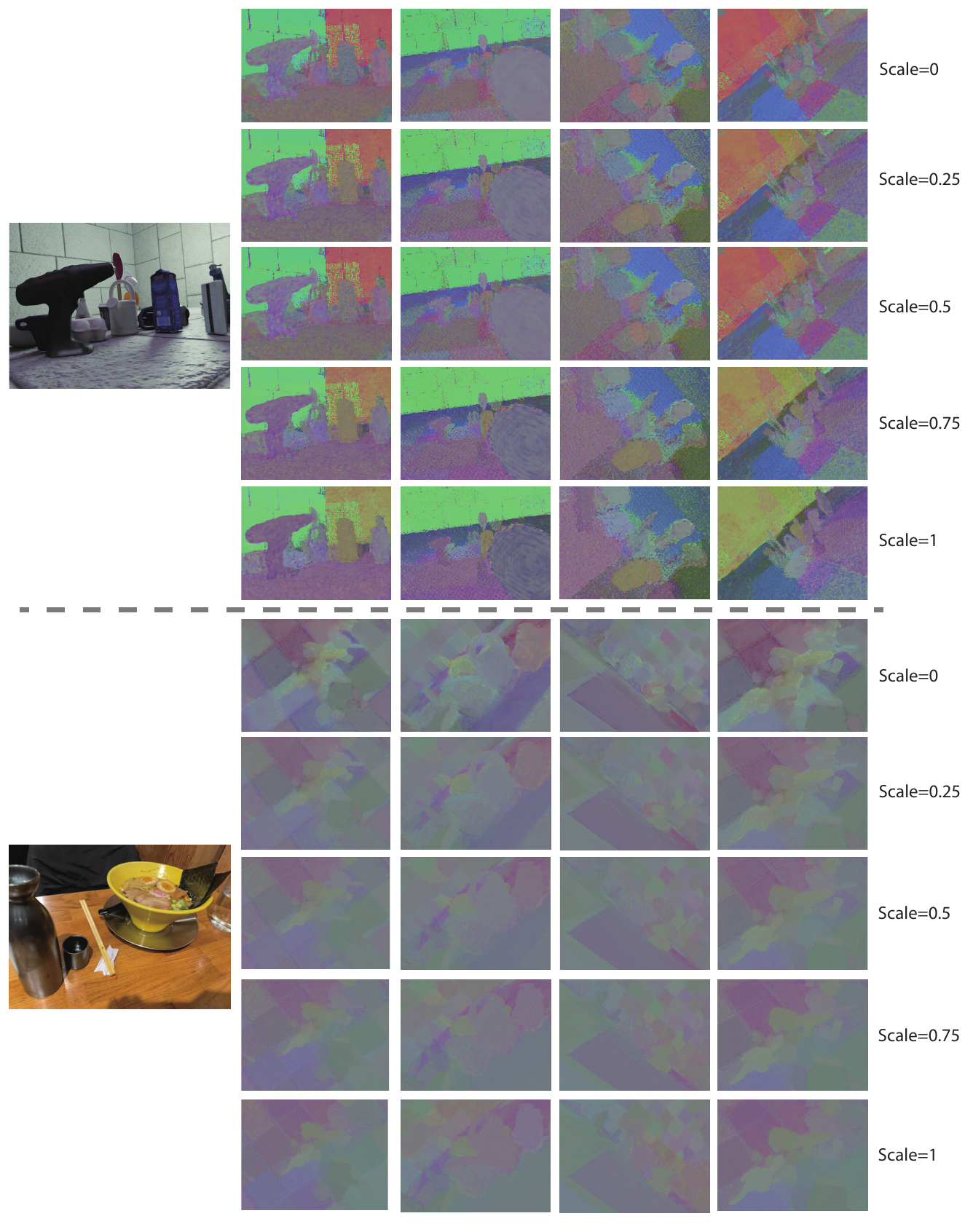}
        \vspace{-0.3in}
	\caption{\textbf{Visualization of hierarchical feature.} The feature space dynamically adjusts its granularity in response to physical scale: finer granularity is employed at smaller scales to capture intricate details, while coarser granularity is utilized at larger scales to emphasize overarching structures. This adaptive mechanism ensures robust and effective representation across a wide range of scales.\label{hrgd} }		
	 \vspace{-0.2in}
\end{figure*}

ReasonGrounder is capable of performing fine-grained Gaussian grouping across multiple scales in diverse scene types, showcasing its versatility in handling complex environments. The hierarchical instance features learned in the Gaussian space are not only compact but also highly effective for decomposing scenes into meaningful components.
Our method employs a hierarchical organization of the entire Gaussian field, structured according to the scale of the queried target object. This hierarchical structuring facilitates the adaptive grouping of Gaussians, ensuring scalability and precision. By isolating the appropriate Gaussian group corresponding to the target, ReasonGrounder enables accurate 3D localization and amodal perception, even for novel viewpoints that were not observed during training.
To provide insights into the learned Gaussian features and their hierarchical structure, visualizations using Principal Component Analysis (PCA) are presented in Figure \ref{hlerf}, \ref{hovs} and \ref{hrgd}. These visualizations illustrate the effectiveness of our method in representing and organizing Gaussian groups at varying scales.

\section{MORE QUANTITATIVE RESULTS}
To complement the mIoU score metric, we also evaluate the 3D-OVS dataset using the Localization Accuracy metric. Accordingly, we compare the performance of our method against other state-of-the-art approaches on this dataset, with the results presented in Table~\ref{table:3dovs_supp}. It is evident from the table that our ReasonGrounder consistently achieves superior performance, further demonstrating the effectiveness and advantages of our approach.

\begin{table}[t]
	\renewcommand\tabcolsep{3pt}
	\centering
	\begin{tabular}{lcccccc}
		\toprule
		Method & \textit{bed}   & \textit{bench} & \textit{room}  & \textit{sofa}  & \textit{lawn}  & \textbf{overall}   \\
		\midrule
		LSeg~\cite{li2022language} & 87.6 & 42.7 & 46.1 & 16.5 & 77.5 & 54.1 \\
		ODISE~\cite{xu2023open} & 86.5 & 39.0 & 59.7 & 35.4 & 82.5 & 60.6 \\
		OV-Seg~\cite{liang2023open} & 40.4 & 89.2 & 49.1 & 69.6 & 92.1 & 68.1 \\
		\midrule
		FFD~\cite{kobayashi2022decomposing} & 86.9 & 42.8 & 51.4 & 9.5 & 82.6 & 54.6 \\
		LERF~\cite{kerr2023lerf} & 86.9 & 79.7 & 79.8 & 43.8 & 93.5 & 76.7 \\
		3D-OVS~\cite{liu2023weakly} & \cellcolor{softyellow}96.7 & \cellcolor{softyellow}96.3 & \cellcolor{softyellow}98.9 & \cellcolor{softyellow}91.6 & \cellcolor{softyellow}97.3 & \cellcolor{softyellow}96.2 \\
		LangSplat~\cite{qin2024langsplat} & \cellcolor{softorange}99.2 & \cellcolor{softorange}98.6 & \cellcolor{softorange}99.3 & \cellcolor{softorange}97.9 & \cellcolor{softorange}99.4 & \cellcolor{softorange}98.9 \\ 
		\textbf{ReasonGrounder} & \cellcolor{softred}\textbf{99.2} & \cellcolor{softred}\textbf{99.1} & \cellcolor{softred}\textbf{99.4} & \cellcolor{softred}\textbf{98.2} & \cellcolor{softred}\textbf{99.4} & \cellcolor{softred}\textbf{99.1} \\ 		
		\bottomrule
	\end{tabular}
       \vspace{-0.1in}
	\caption{\textbf{Localization Accuracy (\%) on 3D-OVS dataset for open-vocabulary 3D visual grounding.} The first three methods target 2D visual grounding, whereas the remaining methods, including our ReasonGrounder, focus on 3D visual grounding.}
	\label{table:3dovs_supp}
	   \vspace{-0.15in}
\end{table}

\section{MORE QUALITATIVE RESULTS}
\begin{figure*}
    \centering
    \includegraphics[width=1\linewidth]{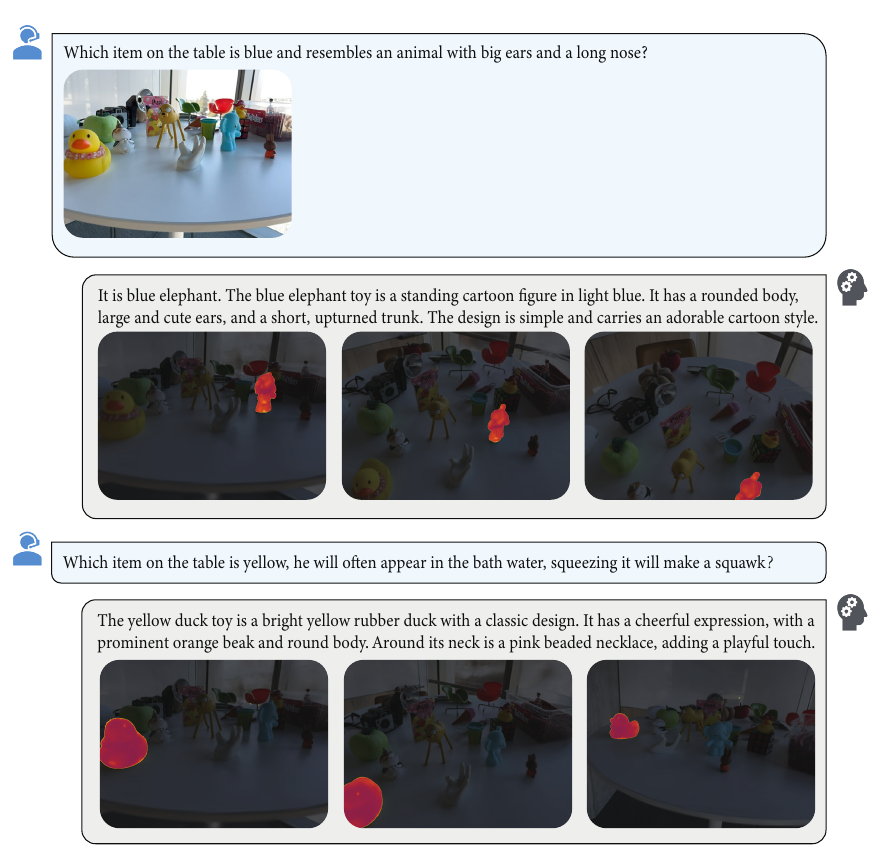}
        \vspace{-0.3in}
	\caption{Qualitative comparisons on the \textbf{\textit{Figurines}} scene of the LERF dataset.\label{qf} }		
	 \vspace{-0.2in}
\end{figure*}

\begin{figure*}
    \centering
    \includegraphics[width=1\linewidth]{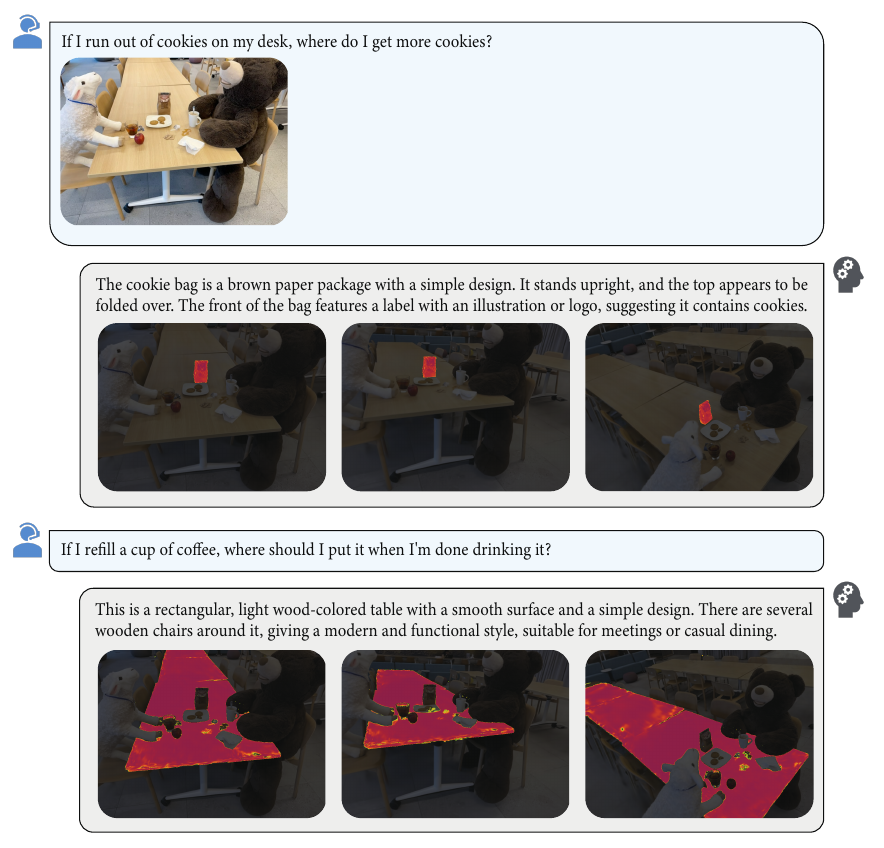}
        \vspace{-0.3in}
	\caption{Qualitative comparisons on the \textbf{\textit{Teatime}} scene of the LERF dataset.\label{qt} }		
	 \vspace{-0.2in}
\end{figure*}

\begin{figure*}
    \centering
    \includegraphics[width=1\linewidth]{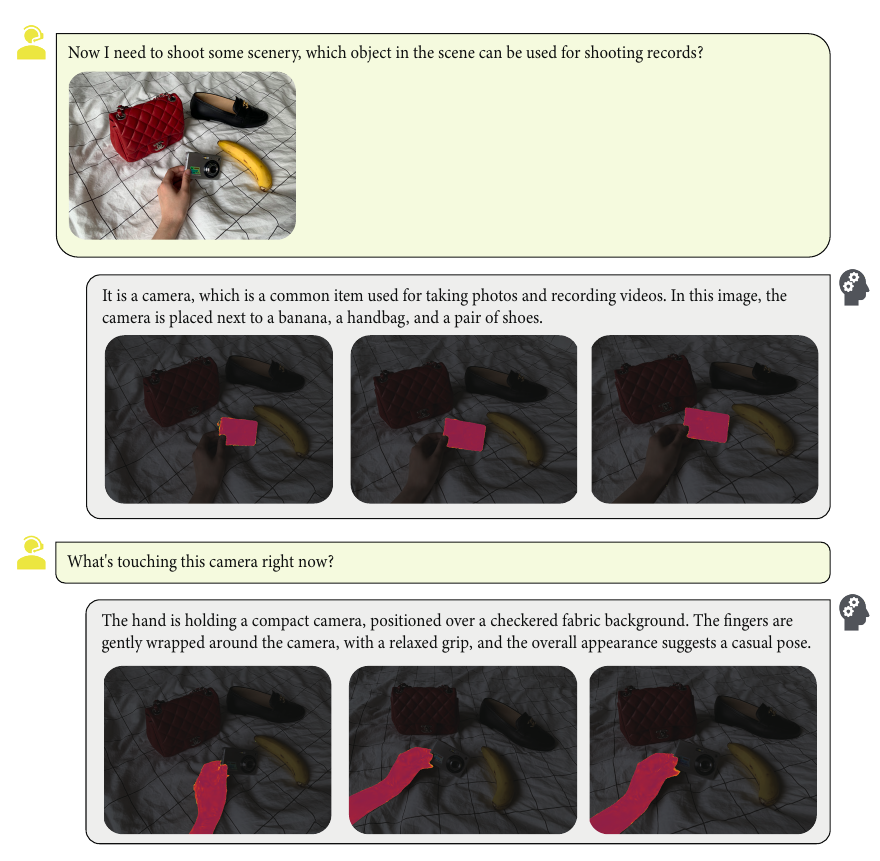}
        \vspace{-0.3in}
	\caption{Qualitative comparisons on the \textbf{\textit{Bed}} scene of the 3D-OVS dataset.\label{qbed} }		
	 \vspace{-0.2in}
\end{figure*}

\begin{figure*}
    \centering
    \includegraphics[width=1\linewidth]{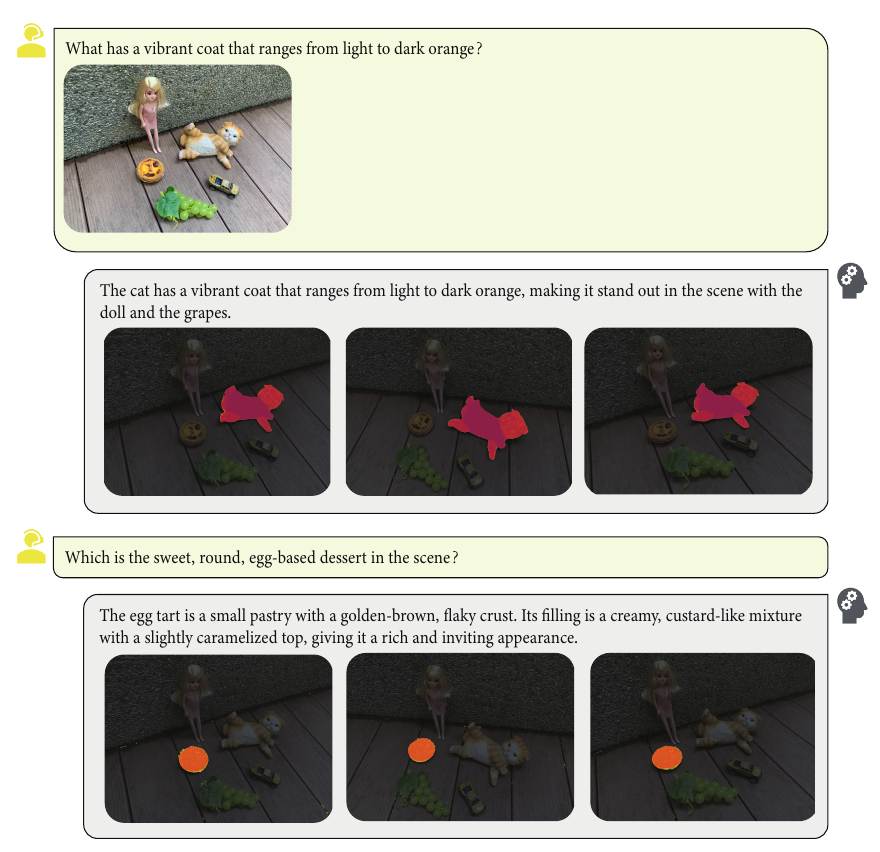}
        \vspace{-0.3in}
	\caption{Qualitative comparisons on the \textbf{\textit{Bench}} scene of the 3D-OVS dataset.\label{qbench} }		
	 \vspace{-0.2in}
\end{figure*}

Qualitative results from scenes not featured in the main text or accompanying videos are presented in Figures \ref{qf}, \ref{qt}, \ref{qbed}, and \ref{qbench}. These additional examples showcase the versatility and robustness of the proposed ReasonGrounder framework across diverse and unseen environments. The experimental results further validate that ReasonGrounder effectively enables open-vocabulary 3D visual grounding and reasoning. Specifically, it demonstrates the capability to accurately localize objects and interpret relationships within complex 3D scenes, even when operating with an open-ended vocabulary. This highlights its potential for addressing a wide range of real-world tasks that require advanced scene understanding and reasoning.


\end{document}